\documentclass{article}

\usepackage[final]{colm2026_conference}

\usepackage{amsmath}
\usepackage{amssymb}
\usepackage{booktabs}
\usepackage{graphicx}
\usepackage{bbding}
\usepackage[dvipsnames,table]{xcolor}
\definecolor{darkblue}{HTML}{0B3D91}
\usepackage{hyperref}
\hypersetup{
  colorlinks=true,
  linkcolor=darkblue,
  citecolor=darkblue,
  urlcolor=darkblue,
  filecolor=darkblue
}
\usepackage{url}
\usepackage{tikz}
\usetikzlibrary{arrows.meta,positioning,calc,shadows.blur}
\usepackage{multirow}
\usepackage{fontawesome5}

\newcommand{\hNA}{\cellcolor{black!8}}      
\newcommand{\hNo}{\cellcolor{red!15}}       
\newcommand{\hWeak}{\cellcolor{darkblue!10}}
\newcommand{\hPart}{\cellcolor{darkblue!25}}
\newcommand{\hFull}{\cellcolor{darkblue!45}}
\newcommand{\hStrong}{\cellcolor{darkblue!62}}

\title{Reasoning over Non-Parametric Memory in Multilingual LLMs: A Cognitive-Based Analysis of Relational Knowledge}

\author{
Zohreh Abbasi Liasi$^{*\dagger}$ \quad Batoul Hammoud$^{*\dagger\ddagger}$ \quad Anas Madkoor$^{\dagger}$ \quad Ehsaneddin Asgari$^{\dagger\S}$ \\
{\normalfont $^{\dagger}$Qatar Computing Research Institute (QCRI), Hamad Bin Khalifa University} \\
{\normalfont $^{\ddagger}$Universitat Pompeu Fabra} \\
{\normalfont {\color{MidnightBlue}\faEnvelope}~\texttt{easgari@hbku.edu.qa}}
}

\begin{document}

\ifcolmsubmission
\linenumbers
\fi

\maketitle

{\renewcommand{\thefootnote}{\fnsymbol{footnote}}%
\footnotetext[1]{Equal contribution.}%
\footnotetext[4]{Corresponding author.}}%
\setcounter{footnote}{0}

\begin{abstract}
Retrieval-augmented generation (RAG) assumes that language models can reason over relations stated in their context rather than fall back on associations memorized during pretraining. We study this evidence-conditioned relational reasoning in isolation, with the facts supplied directly in the prompt rather than retrieved. \textsc{Rel-Bloom} is a controlled multilingual diagnostic of 450 multiple-choice questions built from 30 Wikidata relations; the same source relations are instantiated at five Bloom-inspired levels (Remember, Understand, Apply, Analyze, Evaluate) and under three veracity conditions (Factual, Counterfactual, Neutralized). Across 43 models, six languages, and two multilingual settings (226,800 predictions), three findings stand out. First, accuracy drops sharply from Apply to Analyze in every evaluated open-weight model series, reaching 47.2 percentage points for the Large Granite bucket, but not for either Gemini model. Second, Evaluate items remain a bottleneck even for models near ceiling on the other task types: Gemini~3.1 Pro averages 99.5\% on the other four levels but 60.7\% on Evaluate. Third, item- and relation-level difficulty rankings are consistently preserved across languages, indicating a difficulty structure partly shared across languages within this benchmark. Across knowledge conditions, Counterfactual contexts are generally harder than Factual ones, while removing entity identities can improve higher-level performance in some model regimes. Greater model capability therefore does not by itself ensure reliable reasoning over supplied relational evidence. Relevant resources are available at:
\url{https://huggingface.co/spaces/QCRI/non-parametric-memory-relational-reasoning}.
\end{abstract}

\section{Introduction}

Large language models are increasingly deployed in knowledge-intensive settings, most notably retrieval-augmented generation (RAG), where a model must reason over facts supplied in its context rather than rely solely on information acquired during pretraining \citep{lewis2020rag}. A fundamental challenge in evaluating such systems is separating answers grounded in the supplied evidence from answers driven by memorized associations: a model may answer correctly by drawing on parametric knowledge about a named entity, or answer incorrectly because that knowledge conflicts with the relation stated in context \citep{longpre2021entity}. Existing benchmarks rarely operationalize this distinction directly, so high accuracy does not by itself show whether a model follows the supplied evidence. End-to-end RAG evaluations further confound this question with retrieval quality. We therefore remove retrieval and supply the relevant fact directly in the prompt, measuring accuracy on evidence-conditioned relational questions as a diagnostic of one component on which RAG systems depend.

Bloom's revised taxonomy \citep{krathwohl2002revision} offers a principled framework for characterizing the cognitive complexity of tasks, organizing cognitive processes into six hierarchical levels: Remember, Understand, Apply, Analyze, Evaluate, and Create. Originally designed to classify human learning objectives, the taxonomy has been adopted in NLP and LLM research to evaluate models across varying depths of reasoning \citep{fei2024lawbench,cao2024structeval,abootorabi-etal-2026-almieyar}. A critical assumption underlying such work remains insufficiently tested, however: whether accuracy on Bloom-inspired task types, designed from a taxonomy of human learning objectives, follows the ordering the taxonomy implies. In particular, stronger models may not show a monotonic decline in accuracy across levels, and different level transitions may be associated with different patterns of error.

Multilingual evaluation adds a further dimension to this challenge. Benchmarks such as XTREME \citep{hu2020xtreme} and MEGA \citep{ahuja2023mega} have established cross-lingual performance baselines, yet multilingual evaluations have largely focused on whether models can perform tasks across languages. Less attention has been paid to whether the underlying structure of model behavior itself is stable across languages: whether the same items and relations remain difficult, whether models exhibit similar sensitivity to knowledge format, and whether cognitive difficulty profiles and interference from memorized associations reflect shared model behavior or language-specific artifacts.

We address these gaps simultaneously with \textsc{Rel-Bloom}, a multilingual diagnostic benchmark for evaluating how language models reason over supplied relational knowledge. \textsc{Rel-Bloom} contains 450 multiple-choice questions grounded in 30 curated Wikidata relations \citep{wang2021wikidata5m}. Each question is instantiated under three veracity conditions: Factual, in which real entities are paired with the correct relation; Counterfactual, in which one entity is replaced with an incorrect real entity while preserving the relation; and Neutralized, in which entity identities are replaced with typed abstract placeholders. This design allows us to vary the degree to which parametric entity associations can assist or interfere with reasoning over the supplied context. Questions span five levels adapted from Bloom's revised taxonomy: Remember, Understand, Apply, Analyze, and Evaluate, with Create omitted because it is not naturally compatible with controlled multiple-choice evaluation.

We evaluate 43 models: 41 open-weight models from seven model families, analyzed as eight model series (Qwen2.5 Base, Qwen2.5 Instruct, Qwen3, Qwen3.5, Gemma, LLaMA, Granite, and DeepSeek, with the Qwen2.5 family split into Base and Instruct series), together with two closed-weight Gemini models. Evaluation covers six languages, Arabic, Chinese, English, Hindi, Persian, and Spanish, and two multilingual settings, yielding 226,800 predictions. Our analysis considers model scale, Bloom cognitive level, knowledge veracity, relation usage frequency, and language. Matched accuracy contrasts are evaluated using relation-clustered inference over the 30 source relations, while cross-language stability is assessed using dedicated correlation-based procedures.

\paragraph{Contributions.} (i) \textbf{Benchmark and experimental design.} \textsc{Rel-Bloom} instantiates the same 30 usage-stratified Wikidata relations at five Bloom-inspired levels and under three matched veracity conditions (450 English MCQs), translated into six languages under two evaluation settings, with ten pre-specified hypotheses tested by relation-clustered inference. (ii) \textbf{Findings on Bloom levels and knowledge conditions.} Accuracy drops sharply from Apply to Analyze in every open-weight model series, but not for Gemini, while Evaluate remains a bottleneck even for models near ceiling on the other levels; Counterfactual contexts are generally the hardest condition, and in many open-weight regimes removing entity identities helps more at Analyze and Evaluate than at Remember and Understand. Scaling generally improves accuracy, with series-dependent trajectories. (iii) \textbf{Multilingual stability.} Item- and relation-level difficulty rankings are consistently positively correlated across the six languages. This suggests that \textsc{Rel-Bloom} captures a difficulty structure that is partly shared across languages, despite differences in language-specific performance.
\section{Related Work}

Our work sits at the intersection of three active research threads: cognitive-complexity benchmarking of LLMs, context faithfulness under knowledge conflicts, and the evaluation of reasoning in retrieval-augmented and knowledge-graph settings.

\subsection{Bloom's Taxonomy as a Cognitive Complexity Framework for LLMs}

Bloom's revised taxonomy \citep{krathwohl2002revision} has become a widely adopted scaffold for stratifying the cognitive demands placed on language models, moving evaluation beyond aggregate accuracy toward a diagnosis of where competence breaks down. Early adoption appeared in domain-specific benchmarks such as LawBench \citep{fei2024lawbench} and scientific evaluation suites such as SciEval \citep{sun2024scieval} and SciAssess \citep{cai2024sciassess}. Subsequent work extended this to law \citep{dong2025vlegal}, medicine \citep{zhou2025medbloom}, software engineering \citep{ma2025bloomapr}, and biographical reasoning \citep{cekinmez2025adam}, consistently revealing that performance degrades at higher cognitive levels. StructEval \citep{cao2024structeval} generates test instances automatically for all six levels, while \citet{khoramfar2025deepquestion} report accuracy drops of up to 70\% as tasks ascend the hierarchy, echoing findings from \citet{todorovikj2025kg}, who apply Bloom's framework to characterize tasks in knowledge-graph benchmarks. BloomWise \citep{zoumpoulidi2024bloomwise} reveals a non-monotonic pattern in mathematical reasoning: models score higher on Apply than on Understand; we do not reproduce this ordering in the relational setting.
\citet{raimondi2026mechanistic} further argue, via mechanistic probing, that Bloom's levels correspond to distinguishable internal representations.

\emph{Gap.} None of these benchmarks jointly control (i) matched items across Bloom levels, (ii) factual/counterfactual/entity-neutralized conditions of the same relation, and (iii) multiple languages. \textsc{Rel-Bloom} brings these dimensions together on a single, item-matched design.

\subsection{Context Faithfulness Under Knowledge Conflicts}
A parallel line of work studies whether models ground responses in supplied context or default to memorized knowledge. \citet{longpre2021entity} formalize this as an entity-level knowledge conflict, measuring models' tendency to favor parametric answers over counterfactual context. \citet{li2022controllable} demonstrate that large models exhibit poor controllability in conflict scenarios and introduce Knowledge-Aware Fine-Tuning (KAFT), incorporating counterfactual and irrelevant contexts to improve robustness. \citet{basmov2024llms} show that faithfulness further degrades for hypothetically or modally framed contexts, while \citet{meng2022rome} provide mechanistic evidence linking this behavior to localized factual associations stored in mid-layer feed-forward modules. \citet{liu2025recoglab} introduce a generative framework for relational reasoning that recreates classic cognitive-science paradigms, such as transitive inference and the symbolic distance effect, and report that several human reasoning biases reappear in LLMs. \emph{Gap.} Prior work characterizes when parametric associations override supplied context but rarely varies cognitive demand on the \emph{same} relation. \textsc{Rel-Bloom} adds a Neutralized condition, which removes entity-identity cues without introducing a conflict, and traces how parametric interference scales with Bloom level.

\subsection{RAG Reasoning and Knowledge-Graph QA}

The original RAG framework \citep{lewis2020rag} demonstrated that combining parametric memory with non-parametric retrieval improves factuality on knowledge-intensive tasks. \citet{pan2023unifying} survey the landscape of LLM--KG integration, and GraphRAG variants \citep{thompson2025graphrag} leverage entity-relation graphs for multi-hop traversal and associative retrieval. \citet{sarthi2024raptor} organize documents into hierarchical summaries for improved retrieval. \citet{zhang2025kgqagen} report that the average factual correctness rate across 16 popular KGQA datasets is only 57\%, and \citet{lin2026hybridrag} propose HybridRAG-Bench to control for parametric memorization. \citet{todorovikj2025kg} specifically examine how Bloom's taxonomy can characterize KG tasks, identifying gaps in coverage of higher-order reasoning. \emph{Gap.} Existing RAG/KGQA benchmarks measure whether the retrieved context is used, but not how well the model understands the specific relation carried by that context. \textsc{Rel-Bloom} targets this by evaluating matched relations across factual, counterfactual, and neutralized instantiations of the same triple.

\section{Methods}

\subsection{Benchmark Overview}

We introduce \textsc{Rel-Bloom}, a compact diagnostic benchmark for sentence-level understanding of Wikidata-style relations. Each item consists of a fact sentence, a multiple-choice question with four answer options, and metadata describing the Bloom level, veracity condition, and relation usage tier. \textsc{Rel-Bloom} spans five Bloom-inspired cognitive levels, Remember, Understand, Apply, Analyze, and Evaluate, with 90 questions per level, totalling 450 questions. We omit the sixth level (Create) because it requires open-ended generation that is incompatible with a controlled multiple-choice format. Within each Bloom level, questions are distributed equally across three veracity conditions (Factual, Counterfactual, and Neutralized, 30 questions each), and each set of 30 is further balanced across three relation usage tiers (High, Medium, and Low, 10 questions each).

\begin{figure}[t]
\centering
\begin{tikzpicture}[
  every node/.style={inner sep=0pt, outer sep=0pt},
  box/.style={draw=darkblue!55, line width=0.5pt, rounded corners=3pt,
              minimum width=4.55cm, minimum height=7.3cm,
              align=center, inner xsep=4pt, inner ysep=6pt},
  hdr/.style={font=\footnotesize\bfseries, text=darkblue, align=center},
  key/.style={font=\footnotesize\bfseries, align=center},
  lst/.style={font=\scriptsize, align=center, text=black!85},
  note/.style={font=\scriptsize\itshape, text=darkblue!85!black, align=center},
  arr/.style={-{Latex[length=2.5mm, width=2mm]}, line width=0.7pt, draw=darkblue!70},
  pyrLine/.style={draw=darkblue!55, line width=0.35pt},
]

\node[box] (c1) at (0,0) {};
\node[hdr, anchor=north] (h1) at ([yshift=-3mm]c1.north) {Benchmark construction};

\node[key, anchor=north] (b1) at ([yshift=-3mm]h1.south) {450 English MCQs};
\node[lst, anchor=north] at ([yshift=-1mm]b1.south) {from 30 curated Wikidata relations};

\begin{scope}[shift={(0,-1.05)}, scale=0.72, transform shape]
  \fill[darkblue!6]  (-2.30,0.00) -- (2.30,0.00) -- (1.94,0.55) -- (-1.94,0.55) -- cycle;
  \draw[pyrLine]     (-2.30,0.00) -- (2.30,0.00) -- (1.94,0.55) -- (-1.94,0.55) -- cycle;
  \node[font=\footnotesize] at (0,0.275) {Remember};
  \fill[darkblue!14] (-1.94,0.55) -- (1.94,0.55) -- (1.58,1.10) -- (-1.58,1.10) -- cycle;
  \draw[pyrLine]     (-1.94,0.55) -- (1.94,0.55) -- (1.58,1.10) -- (-1.58,1.10) -- cycle;
  \node[font=\footnotesize] at (0,0.825) {Understand};
  \fill[darkblue!24] (-1.58,1.10) -- (1.58,1.10) -- (1.22,1.65) -- (-1.22,1.65) -- cycle;
  \draw[pyrLine]     (-1.58,1.10) -- (1.58,1.10) -- (1.22,1.65) -- (-1.22,1.65) -- cycle;
  \node[font=\footnotesize] at (0,1.375) {Apply};
  \fill[darkblue!36] (-1.22,1.65) -- (1.22,1.65) -- (0.86,2.20) -- (-0.86,2.20) -- cycle;
  \draw[pyrLine]     (-1.22,1.65) -- (1.22,1.65) -- (0.86,2.20) -- (-0.86,2.20) -- cycle;
  \node[font=\footnotesize] at (0,1.925) {Analyze};
  \fill[darkblue!52] (-0.86,2.20) -- (0.86,2.20) -- (0.50,2.75) -- (-0.50,2.75) -- cycle;
  \draw[pyrLine]     (-0.86,2.20) -- (0.86,2.20) -- (0.50,2.75) -- (-0.50,2.75) -- cycle;
  \node[font=\footnotesize, text=white] at (0,2.475) {Evaluate};
\end{scope}
\node[note, anchor=north] at (0,-1.28) {5 Bloom levels};

\node[key, anchor=north] (b1v) at (0,-1.80) {3 veracity conditions};
\node[lst, anchor=north] at ([yshift=-1mm]b1v.south) {%
  \tikz[baseline=-0.5ex]{%
    \draw[darkblue, line width=0.35pt] (0,0) circle (1.4mm);
    \draw[darkblue, line width=0.55pt, line cap=round]
      (-0.7mm,0.1mm) -- (-0.15mm,-0.55mm) -- (0.85mm,0.65mm);
  }\ Factual, \quad
  \tikz[baseline=-0.5ex]{%
    \draw[darkblue, line width=0.35pt] (0,0) circle (1.4mm);
    \draw[darkblue, line width=0.55pt, line cap=round] (-0.6mm,-0.6mm) -- (0.6mm,0.6mm);
    \draw[darkblue, line width=0.55pt, line cap=round] (-0.6mm,0.6mm) -- (0.6mm,-0.6mm);
  }\ Counterfactual, \\[1pt]
  \tikz[baseline=-0.5ex]{%
    \draw[darkblue, line width=0.4pt, dash pattern=on 0.45mm off 0.35mm] (0,0) circle (1.4mm);
  }\ Neutralized%
};

\node[font=\scriptsize\bfseries, text=darkblue, align=center, fill=darkblue!8,
      rounded corners=2pt, inner xsep=3pt, inner ysep=2.5pt, anchor=south]
  at (0,-3.52) {Same 30 relations at every level \\ and condition: $30 \times 5 \times 3 = 450$};

\node[box, right=6mm of c1] (c2) {};
\node[hdr, anchor=north] (h2) at ([yshift=-3mm]c2.north) {Multilingual evaluation};

\node[key, anchor=north] (b2) at ([yshift=-4mm]h2.south) {6 languages};
\node[lst, anchor=north] at ([yshift=-1mm]b2.south) {Arabic, Chinese, English, \\ Hindi, Persian, Spanish};

\node[key, anchor=north] (b2b) at ([yshift=-22mm]h2.south) {2 evaluation settings};
\node[lst, anchor=north] at ([yshift=-1mm]b2b.south) {Hybrid cross-lingual, \\ Fully translated};

\node[key, anchor=north] (b2c) at ([yshift=-40mm]h2.south) {226{,}800 predictions};
\node[lst, anchor=north] at ([yshift=-1mm]b2c.south) {43 models in fully translated \\ + 41 (open-weight) in hybrid};

\node[box, right=6mm of c2] (c3) {};
\node[hdr, anchor=north] (h3) at ([yshift=-3mm]c3.north) {Models \& analysis};

\node[key, anchor=north] (b3) at ([yshift=-3mm]h3.south) {43 models};

\node[key, anchor=north] (b3a) at ([yshift=-12mm]h3.south) {Open-weight \\ (7 families, 8 series)};
\node[lst, anchor=north] at ([yshift=-1mm]b3a.south) {Qwen2.5 (Base, Instruct), Qwen3, \\ Qwen3.5, Gemma, LLaMA, \\ Granite, DeepSeek};

\node[key, anchor=north] (b3b) at ([yshift=-34mm]h3.south) {Closed-weight (Gemini)};
\node[lst, anchor=north] at ([yshift=-1mm]b3b.south) {Gemini 3 Flash, Gemini 3.1 Pro};

\node[key, anchor=north] (b3c) at ([yshift=-47mm]h3.south) {Analysis dimensions};
\node[lst, anchor=north] at ([yshift=-1mm]b3c.south) {Scale, Bloom level, Veracity, \\ Relation usage frequency, Language};

\draw[arr] (c1.east) -- (c2.west);
\draw[arr] (c2.east) -- (c3.west);

\end{tikzpicture}
\caption{Overview of \textsc{Rel-Bloom}. The same 30 curated Wikidata source relations (10 per relation usage frequency tier) are each instantiated at all five Bloom-inspired levels (shown as a pyramid) and under all three veracity conditions, yielding $30 \times 5 \times 3 = 450$ English items; this full crossing allows level and condition effects to be compared on matched relations. Items are then evaluated in six languages under two multilingual settings. The 43 models comprise 41 open-weight models from seven model families, analyzed as eight model series because Qwen2.5 Base and Qwen2.5 Instruct are treated separately, plus two Gemini models.}
\label{fig:overview}
\end{figure}
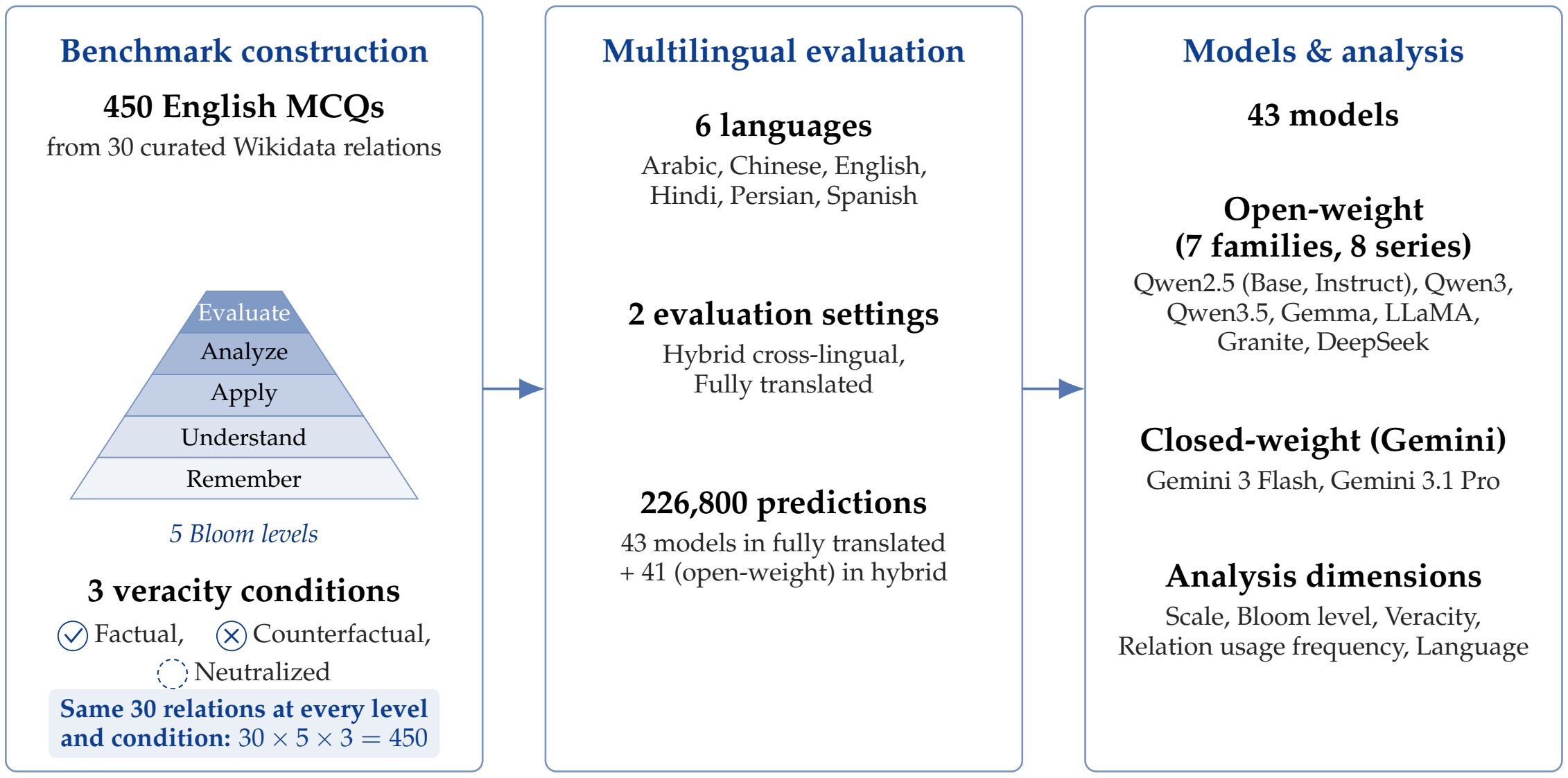

 Figure~\ref{fig:overview} summarizes the design.

\subsection{Relation Selection and Usage Tiers}

We draw candidate relations from Wikidata5M \citep{wang2021wikidata5m} and stratify them into Low, Medium, and High usage tiers using MediaWiki property-frequency counts as a proxy for relation prevalence. These counts approximate how often a relation is used in Wikidata; they are not estimates of how often the relation appears in model pretraining data or of how strongly it is memorized, and we interpret relation usage frequency results (H6) accordingly. We manually curate 10 relations per tier, yielding 30 source relations. The complete sampling, frequency transformation, and stratification procedure is provided in Appendix~\ref{app:construction}. 

\subsection{Triple Extraction and Manual Curation}

For each candidate relation, we extract one representative factual triple $(h,p,t)$ and manually curate a final set of 30 relations, with 10 from each usage tier. Curation prioritizes semantic clarity, domain diversity, and non-overlapping entities, while excluding sensitive or excessively technical relations. Each selected relation contributes one source triple that is reused across Bloom levels and veracity conditions, prioritizing controlled comparison over broad relation coverage. Relation-level results therefore describe these 30 relation instances; effects attributed to a relation may partly reflect the particular entities and fact chosen for it. Further extraction and curation details are provided in Appendix~\ref{app:construction}.

\subsection{Fact Verbalization and Question Generation}
\label{sec:question-gen}

For each extracted triple $(h,p,t)$, GPT-5\footnote{We accessed GPT-5 in early 2026. All verbalized facts were subsequently verified manually by the authors.} verbalizes the fact and assists in generating level-specific multiple-choice questions. Table~\ref{tab:bloom-levels} summarizes the operationalization of the five Bloom levels.

The levels differ in the operation a question requires, and therefore also in how questions and distractors are constructed. Analyze distractors are near-miss relations that can be rejected by checking them against the stated fact, whereas Evaluate distractors are plausible statements about the relation that are not its defining criterion and cannot be rejected in this way. Differences in question construction and distractor difficulty may therefore contribute to accuracy differences between levels. We interpret level-wise gaps as differences in accuracy on these operationalized task types, not as direct measurements of distinct cognitive abilities.

All generated questions, answer options, and answer keys are manually reviewed for factual correctness, preservation of the intended relation, answer uniqueness, and distractor plausibility. As an additional validation pass, Gemini and Grok independently flag disagreements with the provided answers; all flagged cases are manually inspected, and ambiguous items are revised and re-evaluated. Additional quality-control details are provided in Appendix~\ref{app:quality-control}.

\begin{table}[t]
\caption{Bloom-level operationalization in \textsc{Rel-Bloom}.}
\label{tab:bloom-levels}
\centering
\small
\setlength{\tabcolsep}{5pt}
\renewcommand{\arraystretch}{1.05}
\begin{tabular}{@{}lp{11.4cm}@{}}
\toprule
\textbf{Bloom level} & \textbf{Operationalization in \textsc{Rel-Bloom}} \\
\midrule
Remember   & Extract the value explicitly stated in the fact. \\
Understand & Select a paraphrase preserving the relation semantics; distractors express different relations or attributes. \\
Apply      & Apply an abstract relation schema to a new statement with the same subject--relation--object structure and meaning. \\
Analyze    & Discriminate the target relation from near-miss alternatives (semantically related or structurally similar but logically incorrect). \\
Evaluate   & Select the strongest justification grounded in the defining semantic criterion of the relation; distractors are plausible but non-definitional. \\
\bottomrule
\end{tabular}
\end{table}

\subsection{Veracity Conditions}

To separate the contributions of entity-specific prior knowledge, evidence-following, and relational abstraction, we construct three matched veracity conditions for each Bloom level, summarized in Table~\ref{tab:veracity}. The variants preserve the underlying relation and Bloom-level question structure, enabling controlled comparisons across Factual, Counterfactual, and Neutralized conditions. The Counterfactual condition is designed to conflict with knowledge a model is likely to hold about well-known entities; we do not directly measure whether each model has memorized each source fact, so Counterfactual effects indicate sensitivity to likely prior-knowledge conflict rather than verified memorization. 

\begin{table}[t]
\caption{Veracity variants in \textsc{Rel-Bloom}, generated with GPT-5 and manually verified.}
\label{tab:veracity}
\centering
\begin{tabular}{@{}lp{4.2cm}p{4.2cm}@{}}
\toprule
\textbf{Variant} & \textbf{Construction rule} & \textbf{What it tests} \\
\midrule
Factual & Verbalize the extracted triple $(h, p, t)$ as a sentence-level fact. & Baseline relation understanding with real entity surface forms. \\
Counterfactual & Edit exactly one argument of $(h, p, t)$, preserving coarse entity type and relation context. & Evidence-following under conflict with parametric priors. \\
Neutralized & Replace entity mentions with typed placeholders (e.g., \texttt{[PERSON~1]}, \texttt{[SPORT~1]}) while preserving the question template. & Relation abstraction with reduced entity-prior shortcuts. \\
\bottomrule
\end{tabular}
\end{table}

\subsection{Multilingual Evaluation Settings}

\textsc{Rel-Bloom} is constructed in English and evaluated in Arabic, Chinese, English, Hindi, Persian, and Spanish under two settings. In the \emph{hybrid cross-lingual setting}, the fact and question remain in English while the answer options are translated into the target language, isolating cross-lingual answer selection from a shared English stimulus. In the \emph{fully translated setting}, the fact, question, and options are all presented in the target language while preserving the relation, Bloom-level operation, veracity condition, answer key, and option order. We analyze the two settings separately.

\paragraph{Reproducibility.} Model checkpoints (Appendix~\ref{app:model-inventory}), benchmark construction (Appendix~\ref{app:construction}), quality control (Appendix~\ref{app:quality-control}), and hypothesis-specific test statistics, permutation/cluster-bootstrap procedures, and the Holm correction family (Appendix~\ref{app:hypotheses}) are documented. The complete benchmark, translations, prompt templates, decoding configurations, and per-model prediction logs will be released upon publication.

\subsection{Hypothesis Testing Framework}

\paragraph{Hypothesis catalog.}
We evaluate ten pre-specified (not publicly preregistered) hypotheses spanning model scale, Bloom-level difficulty, veracity, relation usage frequency, and cross-language stability, listed in Table~\ref{tab:hypotheses}. For scale-dependent analyses, open-weight models are grouped into Small ($<7$B), Medium ($\geq7$B and $<14$B), and Large ($\geq14$B) buckets; Gemini models are reported individually because their parameter counts are undisclosed. Checkpoints within a bucket and across buckets also differ in training data, recipe, and in some cases architecture (Appendix~\ref{app:model-inventory}), so size-bucket comparisons are descriptive rather than causal estimates of scaling effects.

\begin{table}[h]
\caption{The ten hypotheses tested in \textsc{Rel-Bloom}, grouped by theme.}
\label{tab:hypotheses}
\centering
\small
\setlength{\tabcolsep}{5pt}
\renewcommand{\arraystretch}{1.15}
\begin{tabular}{@{}l l p{9.4cm}@{}}
\toprule
\textbf{Theme} & \textbf{ID} & \textbf{Statement} \\
\midrule
\multirow{2}{*}{\textcolor{darkblue}{\textbf{Scale}}}
  & \textbf{H1} & Diminishing returns: Small$\to$Medium gain exceeds Medium$\to$Large. \\
  & \textbf{H2} & Monotonic scaling: Medium $>$ Small \emph{and} Large $>$ Medium. \\
\midrule
\multirow{2}{*}{\textcolor{darkblue}{\textbf{Bloom level}}}
  & \textbf{H3} & Apply $>$ Analyze accuracy; a ``cliff'' is marked when the lower 95\% CI on the drop exceeds 10\,pp. \\
  & \textbf{H8} & Evaluate underperforms the mean of the other four Bloom levels. \\
\midrule
\multirow{3}{*}{\textcolor{darkblue}{\textbf{Veracity}}}
  & \textbf{H4} & Ordering: Neutralized $>$ Factual $>$ Counterfactual. \\
  & \textbf{H5} & Factual--Counterfactual gap is larger at Analyze/Evaluate than at Remember/Understand. \\
  & \textbf{H7} & Neutralized-over-Factual advantage is larger at Analyze/Evaluate than at Remember/Understand. \\
\midrule
\textcolor{darkblue}{\textbf{Usage}}
  & \textbf{H6} & Relation usage frequency has a practically negligible effect (observed Low/Medium/High spread $<5$\,pp; equivalence assessed within a $\pm5$\,pp margin). \\
\midrule
\multirow{2}{*}{\textcolor{darkblue}{\textbf{Language}}}
  & \textbf{H9} & Item-level difficulty rankings are preserved across the six languages (mean of 15 pairwise Spearman correlations). \\
  & \textbf{H10} & Relation-level difficulty rankings are preserved across languages (relation-identity permutation test). \\
\bottomrule
\end{tabular}
\end{table}

\paragraph{Estimation and inference.}
For H1--H5, H7, and H8, matched contrasts are aggregated within the 30 source relations and evaluated using one-sided relation-cluster sign-flip tests with 95\% relation-cluster bootstrap intervals. H6 is evaluated using an observed Low/Medium/High spread below 5\,pp together with a $\pm5$\,pp equivalence margin for the three pairwise usage contrasts; Medium and Large are the confirmatory buckets, while Small is reported exploratorily. For cross-language stability, H9 summarizes the 15 pairwise item-level Spearman correlations and evaluates their mean using a one-sided sign-flip test, whereas H10 uses a relation-identity permutation test in which relation correspondences are shuffled across languages and the full mean-correlation statistic is recomputed. Holm correction is applied within hypothesis-specific families, typically the S/M/L tests within each series; full definitions and correction families are given in Appendix~\ref{app:hypotheses}.

\section{Results}

Table~\ref{tab:support-summary} summarizes hypothesis support across the fully translated and hybrid cross-lingual settings; the full per-series, per-bucket matrices are given in Appendix~\ref{app:support-matrices} (Tables~\ref{tab:support-ft} and~\ref{tab:support-hy}). The most consistent findings are cross-language stability (\textbf{\underline{H9}}--\textbf{\underline{H10}}), the Apply-to-Analyze drop in open-weight models (\textbf{\underline{H3}}), the Evaluate bottleneck (\textbf{\underline{H8}}), and Small-to-Medium scaling (\textbf{\underline{H2}}); \textbf{\underline{H1}} and \textbf{\underline{H4}}--\textbf{\underline{H7}} depend more on model series and size bucket.

\begin{table}[t]
\caption{Summary of hypothesis support (Holm-corrected; full matrices in Tables~\ref{tab:support-ft}--\ref{tab:support-hy}). Units differ between blocks and are stated in the block headers and cells. The Gemini column counts supported models (fully translated setting only).}
\label{tab:support-summary}
\centering
\footnotesize
\setlength{\tabcolsep}{4pt}
\renewcommand{\arraystretch}{1.1}
\begin{tabular}{@{}l l c c c@{}}
\toprule
 & & \multicolumn{2}{c}{\textbf{Open-weight series (of 8)}} & \textbf{Gemini (of 2)} \\
\cmidrule(lr){3-4}\cmidrule(l){5-5}
\textbf{ID} & \textbf{Hypothesis} & \textbf{Fully transl.} & \textbf{Hybrid} & \textbf{Fully transl.} \\
\midrule
\multicolumn{5}{@{}l}{\emph{Per-bucket tests: series significant in all 3 buckets (in $\geq$1 bucket)}} \\
H9  & Item-level cross-language stability     & 8 (8) & 8 (8) & 2 \\
H10 & Relation-level cross-language stability & 8 (8) & 8 (8) & 2 \\
H3  & Apply $>$ Analyze                       & 7 (8) & 6 (8) & 0 \\
H8  & Evaluate bottleneck                     & 6 (8) & 5 (8) & 2 \\
H7  & Neutralized gain larger at high levels & 3 (8) & 6 (8) & 0 \\
H5  & F--CF gap larger at high levels         & 0 (1) & 0 (2) & 2 \\
H4  & Neutralized $>$ Factual $>$ Counterfactual & 0 (1) & 0 (5) & 0 \\
\midrule
\multicolumn{5}{@{}l}{\emph{Other units}} \\
H2  & Monotonic scaling (per transition)      & M$>$S: 8, L$>$M: 7 & M$>$S: 8, L$>$M: 7 & \textemdash{} \\
H1  & Diminishing returns (series-level test) & 5 series & 6 series & \textemdash{} \\
H6  & Negligible usage effect (equivalence)   & 0 series & 0 series & 0 \\
\bottomrule
\end{tabular}
\end{table}

\subsection{Scaling Behavior}

Accuracy generally increases with model size (\textbf{\underline{H2}}). In the fully translated setting, Medium significantly exceeds Small in all eight open-weight series, while Large significantly exceeds Medium in seven; DeepSeek is the only exception. The same broad pattern holds in the hybrid cross-lingual setting. For example, Qwen2.5 Instruct gains 23.8 percentage points from Small to Medium and 6.1 points from Medium to Large.

Diminishing returns (\textbf{\underline{H1}}) are less universal, holding in five of eight series in the fully translated setting and six of eight in the hybrid cross-lingual setting. LLaMA and Granite show larger upper-transition gains, while DeepSeek supports (\textbf{\underline{H1}}) only in the hybrid cross-lingual setting and slightly declines from Medium to Large when fully translated. Qwen2.5 Base and Instruct show similar scaling trends.

\subsection{Bloom-Level Patterns}

Apply exceeds Analyze across all eight open-weight series in both settings (\textbf{\underline{H3}}), but the magnitude of the drop varies sharply with series and size bucket. A ``cliff'' denotes a statistically supported drop whose lower 95\% relation-cluster bootstrap bound exceeds 10 percentage points. At Large, Granite exhibits a 47.2-point drop, whereas Qwen3.5 drops only 7.2 points. Neither Gemini model exhibits the Apply-to-Analyze accuracy drop observed across the evaluated open-weight model series. Across open-weight series, the cliff is neither scale-invariant nor monotonic with model size.

Evaluate presents a different pattern (\textbf{\underline{H8}}). Gemini~3.1 Pro averages approximately 99.5\% across the other four levels but only 60.7\% on Evaluate, a gap of about 39 percentage points for a model that shows no Apply-to-Analyze drop at all. The Evaluate gap appears in all eight open-weight series and both Gemini models (Table~\ref{tab:support-summary}). The two task types thus show different model-dependent accuracy patterns (\S\ref{sec:discussion}); as noted in \S\ref{sec:question-gen}, these gaps are differences between operationalized task types and may partly reflect item and distractor construction.

\subsection{Veracity Effects}

The full Neutralized $>$ Factual $>$ Counterfactual ordering (\textbf{\underline{H4}}) is common directionally but statistically supported in only a few size buckets and in no series as a whole. Counterfactual is generally the weakest condition, whereas Neutralized does not consistently exceed Factual and reverses in several high-capability models, including Qwen3.5, Gemma, and both Gemini models. 

The Factual--Counterfactual gap (\textbf{\underline{H5}}) also depends on cognitive demand. Its expected widening from lower to higher Bloom levels appears directionally in several open-weight series but is statistically supported in only a few size buckets. The effect is strongest for Gemini, where both models show significant widening despite near-ceiling performance on the lower-complexity levels. 

\subsection{Neutralization and Cognitive Demand}

\textbf{\underline{H7}} compares the Neutralized-over-Factual advantage at Analyze and Evaluate with that at Remember and Understand. In the fully translated setting, the high-complexity advantage is significantly larger in the Small bucket for seven of eight open-weight series and remains present in several Medium and Large regimes.

The effect weakens or reverses in some stronger series: it remains substantial for Qwen2.5 Base and Instruct, becomes smaller for Qwen3, and is absent or reversed for several of the strongest models, including Gemini~3.1 Pro. Support for \textbf{\underline{H7}} is therefore recurrent but not universal, varying across model series and size buckets.

\subsection{Relation Usage Frequency}

\textbf{\underline{H6}} receives limited support. Although several Medium and Large buckets exhibit observed Low/Medium/High spreads below five percentage points, no open-weight series satisfies the stronger interval-based equivalence criterion in the fully translated setting. Gemini~3.1 Pro also shows a small observed spread, whereas Gemini~3 Flash lies at the five-point boundary under the rounded result. We therefore regard the influence of relation usage frequency as inconclusive rather than absent. Because the tiers reflect Wikidata property counts, this result says nothing about pretraining exposure or memorization strength.

\subsection{Cross-Language Stability}

Item-level difficulty rankings (\textbf{\underline{H9}}) show consistently positive cross-language stability across every model series and size bucket in both settings and for both Gemini models. Significance alone does not convey strength: in the fully translated setting, the mean pairwise correlation is approximately $\rho=0.54$ for Gemini~3 Flash, a moderate level of agreement, and $\rho=0.74$ for Gemini~3.1 Pro, a strong one.

Relation-level difficulty rankings (\textbf{\underline{H10}}) show the same pattern: all series, size buckets, and both Gemini models exhibit significantly positive cross-language stability. Thus, differences in aggregate accuracy coexist with a difficulty structure that is partly shared across languages. Because all language versions derive from the same English items and translated templates, shared item construction may contribute to these correlations, as may the single triple per relation. The results establish stability for the evaluated relation instances, not language-independent reasoning ability in general.

\section{Analysis}
\label{sec:discussion}

\subsection{How Analyze and Evaluate Items Differ}
\label{sec:analyze-evaluate}

The two task types show different model-dependent accuracy patterns. The Apply-to-Analyze drop varies with series and size and is absent for Gemini, whereas the Evaluate gap persists for every series and both Gemini models. Their difficulty is therefore not captured by a single smooth progression across levels, although this alone does not show that they tap different underlying abilities: a harder task can remain hard after an easier one saturates.

The operationalization in Table~\ref{tab:bloom-levels} offers one account that is consistent with this pattern, though our experiments do not isolate it. Analyze distractors are near-miss relations that are inconsistent with the stated fact, so an item can in principle be solved by checking each option against the evidence. Evaluate distractors are consistent with the fact but are not the defining criterion of the relation, so such checking does not eliminate them.

An equally important alternative is distractor plausibility: Evaluate distractors are by construction more plausible, and the gap may reflect discrimination among harder alternatives rather than a distinct operation. Separating these accounts requires items that vary distractor plausibility within a fixed level. ``Definitional'' criteria are also partly conventional, so some Evaluate errors may reflect reasonable disagreement with the key despite manual review and independent model cross-checks (Appendix~\ref{app:quality-control}). We therefore report the Analyze--Evaluate contrast as a difference between operationalized task types, not between cognitive capabilities.

\subsection{The Effect of Entity Cues Depends on Task Complexity}
\label{sec:entity-crossover}

Averaged over Remember and Understand, Neutralized contexts usually underperform Factual contexts across model-series and size-bucket combinations, suggesting that entity names can provide useful cues for simpler tasks. At Analyze and Evaluate, the pattern often reverses, consistent with entity-specific associations becoming more disruptive as relational demands increase.

This crossover underlies H7. One possibility is that familiar entities help locate the stated fact at lower levels but supply competing associations when the task concerns the relation itself. The Neutralized advantage shrinks in several higher-performing series, consistent with stronger models being less affected by such competition. Because this comparison is exploratory and rests on one source triple for each of 30 relations, we neither interpret it as a scaling law nor claim it generalizes to other triples of the same relations.

\subsection{Errors Consistent with Memory Competition}

We inspect matched Neutralized--Factual pairs for which Qwen2.5-32B-Instruct succeeds under Neutralized input but fails under Factual input. The Factual errors often select a salient attribute or related fact associated with the named entity rather than the relation requested by the question. Table~\ref{tab:memory-competition} shows representative examples. Although these cases do not establish an internal causal mechanism, they are consistent with entity-specific parametric associations competing with the relational evidence supplied in context.

\begin{table}[t]
\caption{Representative errors consistent with memory competition. In the Factual condition, the model selects a salient entity-associated answer rather than the relation required by the question.}
\label{tab:memory-competition}
\centering
\begin{tabular}{@{}lll@{}}
\toprule
\textbf{Entity} & \textbf{Wrong answer} & \textbf{Correct relational answer} \\
\midrule
Apollo 11 & Returned safely to Earth & Recovered by USS Hornet \\
Bohemian Rhapsody & A six-minute song & Tonality is B-flat major \\
Amanita muscaria & Red cap with white spots & Spore print color is white \\
\bottomrule
\end{tabular}
\end{table}

\section{Conclusions}

We presented \textsc{Rel-Bloom}, a controlled multilingual diagnostic of evidence-conditioned relational reasoning over matched relations, Bloom-inspired task types, and veracity conditions. Its results identify limitations in evidence-conditioned relational reasoning that may affect the reliability of downstream RAG systems.

They suggest three concrete practices for evaluating RAG systems. First, report accuracy separately by the operation a question requires: a system that reliably checks retrieved relations against the question (Analyze) may still fail to identify which retrieved property justifies an answer (Evaluate), and aggregate accuracy hides this. Second, include counterfactual and entity-neutralized variants of retrieved passages, since accuracy on factual passages alone conflates evidence-following with prior knowledge about the entities involved. Third, since difficulty rankings transfer across languages here, a translated item set can serve as a first multilingual screen, though language-specific accuracy must still be measured directly.

\paragraph{Limitations and future work.} \textsc{Rel-Bloom} favors controlled comparison over breadth. With one source triple per relation, the 30 relation clusters give limited power for small effects (notably H6) and for generalizing to unseen triples of the same relations. We do not measure end-to-end RAG with noisy retrieval or open-ended generation. Level-wise gaps may reflect item construction; memorization is not measured; size comparisons are descriptive; shared templates may inflate cross-language correlations; and the error analysis uses one checkpoint. Future work can add triples per relation and vary distractor plausibility within levels.
\section*{Ethics Statement}                                                                                                              
In accordance with COLM 2026 policy, we disclose that \textbf{Claude}, \textbf{GPT-5}, and \textbf{Gemini} were used to assist in        
rewriting and improving manuscript drafts across revision rounds. All scientific content, experimental design, and conclusions are the  authors' own responsibility. 



\bibliography{colm2026_conference}

@article{krathwohl2002revision,
  author = {Krathwohl, David R.},
  title = {A Revision of {B}loom's Taxonomy: An Overview},
  journal = {Theory Into Practice},
  volume = {41},
  number = {4},
  pages = {212--218},
  year = {2002},
  doi = {10.1207/s15430421tip4104_2},
  publisher = {Informa UK Limited},
}

@inproceedings{fei2024lawbench,
  title     = {{LawBench}: Benchmarking Legal Knowledge of Large Language Models},
  author    = {Fei, Zhiwei and Shen, Xiaoyu and Zhu, Dawei and
               Zhou, Fengzhe and Han, Zhuo and Huang, Alan and
               Zhang, Songyang and Chen, Kai and Yin, Zhixin and
               Shen, Zongwen and Ge, Jidong and Ng, Vincent},
  booktitle = {Proceedings of the 2024 Conference on Empirical
               Methods in Natural Language Processing},
  pages     = {7933--7962},
  year      = {2024},
  publisher = {Association for Computational Linguistics},
  doi       = {10.18653/v1/2024.emnlp-main.452},
  url       = {https://aclanthology.org/2024.emnlp-main.452/}
}

@inproceedings{abootorabi-etal-2026-almieyar,
    title = "Almieyar-Oryx-{B}loom{B}ench: A Bilingual Multimodal Benchmark for Cognitively Informed Evaluation of Vision-Language Models",
    author = "Abootorabi, Mohammad Mahdi  and
      Ghahroodi, Omid  and
      Madkoor, Anas  and
      Nouri, Marzia  and
      Dastgheib, Doratossadat  and
      Asgari, Ehsaneddin",
    editor = "Liakata, Maria  and
      Moreira, Viviane P.  and
      Zhang, Jiajun  and
      Jurgens, David",
    booktitle = "Findings of the {A}ssociation for {C}omputational {L}inguistics: {ACL} 2026",
    month = jul,
    year = "2026",
    address = "San Diego, California, United States",
    publisher = "Association for Computational Linguistics",
    url = "https://aclanthology.org/2026.findings-acl.1416/",
    doi = "10.18653/v1/2026.findings-acl.1416",
    pages = "28404--28436",
    ISBN = "979-8-89176-395-1",
}

@article{sun2024scieval,
  title   = {{SciEval}: A Multi-Level Large Language Model
             Evaluation Benchmark for Scientific Research},
  author  = {Sun, Liangtai and Han, Yang and Zhao, Zihan and
             Ma, Da and Shen, Zhennan and Chen, Baocai and
             Chen, Lu and Yu, Kai},
  journal = {Proceedings of the AAAI Conference on
             Artificial Intelligence},
  volume  = {38},
  number  = {17},
  pages   = {19053--19061},
  year    = {2024},
  doi     = {10.1609/aaai.v38i17.29872},
  url     = {https://ojs.aaai.org/index.php/AAAI/article/view/29872}
}

@inproceedings{cai2024sciassess,
  author = {Cai, Hengxing and Cai, Xiaochen and Chang, Junhan and Li, Sihang and Yao, Lin and Changxin, Wang and Gao, Zhifeng and Wang, Hongshuai and Yongge, Li and Lin, Mujie and others},
  title = {{SciAssess}: Benchmarking {LLM} Proficiency in Scientific Literature Analysis},
  booktitle = {Findings of the Association for Computational Linguistics: NAACL 2025},
  year = {2025},
  pages = {2335--2357},
  doi = {10.18653/v1/2025.findings-naacl.125},
  publisher = {Association for Computational Linguistics},
}

@inproceedings{cao2024structeval,
  author = {Cao, Boxi and Ren, Mengjie and Lin, Hongyu and Han, Xianpei and Zhang, Feng and Zhan, Junfeng and Sun, Le},
  title = {{StructEval}: Deepen and Broaden Large Language Model Assessment via Structured Evaluation},
  booktitle = {Findings of the Association for Computational Linguistics: ACL 2024},
  year = {2024},
  pages = {5300--5318},
  doi = {10.18653/v1/2024.findings-acl.314},
  publisher = {Association for Computational Linguistics},
}

@inproceedings{khoramfar2025deepquestion,
  author = {Khoramfar, Ali and Ramezani, Ali and Mohajeri, Mohammad Mahdi and Dousti, Mohammad Javad and Ahmadabadi, Majid Nili and Faili, Heshaam},
  title = {{DEEPQUESTION}: Systematic Generation of Real-World Challenges for Evaluating {LLM}s Performance},
  booktitle = {Proceedings of the Language Resources and Evaluation Conference},
  year = {2026},
  pages = {11451--11460},
  publisher = {European Language Resources Association (ELRA)},
}

@inproceedings{zoumpoulidi2024bloomwise,
  author = {Zoumpoulidi, Maria-Eleni and Paraskevopoulos, Georgios and Potamianos, Alexandros},
  title = {{BloomWise}: Enhancing Problem-Solving Capabilities of Large Language Models Using {B}loom's-Taxonomy-Inspired Prompts},
  booktitle = {Proceedings of The 3rd Workshop on Mathematical Natural Language Processing (MathNLP 2025)},
  year = {2025},
  pages = {34--49},
  doi = {10.18653/v1/2025.mathnlp-main.3},
  publisher = {Association for Computational Linguistics},
}

@article{dong2025vlegal,
  author = {Dong, Nguyen Tien and Nguyen, Minh-Anh and Hoang, Thanh Dat and Ngoc, Nguyen Tuan and Minh, Dao Xuan Quang and Hai, Phan Phi and Anh, Nguyen Thi Ngoc and Tu, Dang Van and Vu, Binh},
  title = {{VLegal-Bench}: Cognitively Grounded Benchmark for Vietnamese Legal Reasoning of Large Language Models},
  journal = {arXiv preprint arXiv:2512.14554},
  year = {2025},
  doi = {10.48550/arXiv.2512.14554},
}

@article{ma2025bloomapr,
  title = {{BloomAPR}: A Bloom's Taxonomy-based Framework
           for Assessing the Capabilities of
           {LLM}-Powered {APR} Solutions},
  author = {Ma, Yinghang and
            Shin, Jiho and
            da Silva, Leuson and
            Jiang, Zhen Ming and
            Wang, Song and
            Khomh, Foutse and
            Tan, Shin Hwei},
  journal = {arXiv preprint arXiv:2509.25465},
  year = {2025},
  doi = {10.48550/arXiv.2509.25465},
  url = {https://arxiv.org/abs/2509.25465}
}

@article{cekinmez2025adam,
  author = {Cekinmez, Jasin and Ghahroodi, Omid and Chandle, Saad Fowad and Gupta, Dhiman and Asgari, Ehsaneddin},
  title = {{ADAM}: A Diverse Archive of Mankind for Evaluating and Enhancing {LLM}s in Biographical Reasoning},
  journal = {arXiv preprint arXiv:2509.22991},
  year = {2025},
  doi = {10.48550/arXiv.2509.22991},
}

@inproceedings{todorovikj2025kg,
  author = {Todorovikj, Sara and Meyer, Lars-Peter and Martin, Michael},
  title = {Characterizing Knowledge Graph Tasks in {LLM} Benchmarks Using Cognitive Complexity Frameworks},
  booktitle = {Proceedings of SEMANTiCS 2025 Poster Track},
  year = {2025},
  doi = {10.48550/arXiv.2509.19347},
}

@inproceedings{zhou2025medbloom,
  author = {Zhou, Yuxuan and Liu, Xien and Yan, Chenwei and Ning, Chen and Zhang, Xiao and Li, Boxun and Fu, Xiangling and Wang, Shijin and Hu, Guoping and Wang, Yu and others},
  title = {Evaluating {LLM}s Across Multi-Cognitive Levels: From Medical Knowledge Mastery to Scenario-Based Problem Solving},
  booktitle = {International Conference on Machine Learning},
  year = {2025},
  doi = {10.48550/arXiv.2506.08349},
}

@article{raimondi2026mechanistic,
  author = {Raimondi, Bianca and Gabbrielli, Maurizio},
  title = {Mechanistic Interpretability of Cognitive Complexity in {LLM}s via Linear Probing using {B}loom's Taxonomy},
  journal = {arXiv preprint arXiv:2602.17229},
  year = {2026},
  doi = {10.48550/arXiv.2602.17229},
}

@inproceedings{longpre2021entity,
  author = {Longpre, Shayne and Perisetla, Kartik and Chen, Anthony and Ramesh, Nikhil and DuBois, Chris and Singh, Sameer},
  title = {Entity-Based Knowledge Conflicts in Question Answering},
  booktitle = {Conference on Empirical Methods in Natural Language Processing},
  pages = {7052--7063},
  year = {2021},
  doi = {10.18653/v1/2021.emnlp-main.565},
  publisher = {Association for Computational Linguistics},
}

@inproceedings{li2022controllable,
  author = {Li, Daliang and Rawat, Ankit Singh and Zaheer, Manzil and Wang, Xin and Lukasik, Michal and Veit, Andreas and Yu, Felix X. and Kumar, Sanjiv},
  title = {Large Language Models with Controllable Working Memory},
  booktitle = {Findings of the Association for Computational Linguistics: ACL 2023},
  year = {2023},
  pages = {1774--1793},
  doi = {10.18653/v1/2023.findings-acl.112},
  publisher = {Association for Computational Linguistics},
}

@article{basmov2024llms,
  author = {Basmov, Victoria and Goldberg, Yoav and Tsarfaty, Reut},
  title = {{LLM}s' Reading Comprehension Is Affected by Parametric Knowledge and Struggles with Hypothetical Statements},
  journal = {arXiv preprint arXiv:2404.06283},
  year = {2024},
  doi = {10.48550/arXiv.2404.06283},
}

@inproceedings{meng2022rome,
  author = {Meng, Kevin and Bau, David and Andonian, Alex and Belinkov, Yonatan},
  title = {Locating and Editing Factual Associations in {GPT}},
  booktitle = {Advances in Neural Information Processing Systems},
  volume = {35},
  year = {2022},
  pages = {17359--17372},
}

@inproceedings{lewis2020rag,
  author = {Lewis, Patrick and Perez, Ethan and Piktus, Aleksandra and Petroni, Fabio and Karpukhin, Vladimir and Goyal, Naman and K{\"u}ttler, Heinrich and Lewis, Mike and Yih, Wen-tau and Rockt{\"a}schel, Tim and others},
  title = {Retrieval-Augmented Generation for Knowledge-Intensive {NLP} Tasks},
  booktitle = {Advances in Neural Information Processing Systems},
  volume = {33},
  pages = {9459--9474},
  year = {2020},
}

@article{pan2023unifying,
  author = {Pan, Shirui and Luo, Linhao and Wang, Yufei and Chen, Chen and Wang, Jiapu and Wu, Xindong},
  title = {Unifying Large Language Models and Knowledge Graphs: A Roadmap},
  journal = {IEEE Transactions on Knowledge and Data Engineering},
  year = {2024},
  volume = {36},
  number = {7},
  pages = {3580--3599},
  doi = {10.1109/TKDE.2024.3352100},
  publisher = {Institute of Electrical and Electronics Engineers (IEEE)},
}

@article{thompson2025graphrag,
  author = {Thompson, Travis and Lim, Seung-Hwan and Liu, Paul and He, Ruoying and Xu, Dongkuan},
  title = {Inference Scaled {GraphRAG}: Improving Multi Hop Question Answering on Knowledge Graphs},
  journal = {arXiv preprint arXiv:2506.19967},
  year = {2025},
  doi = {10.48550/arXiv.2506.19967},
}

@inproceedings{sarthi2024raptor,
  author = {Sarthi, Parth and Abdullah, Salman and Tuli, Aditi and Khanna, Shubh and Goldie, Anna and Manning, Christopher D.},
  title = {{RAPTOR}: Recursive Abstractive Processing for Tree-Organized Retrieval},
  booktitle = {The Twelfth International Conference on Learning Representations},
  year = {2024},
}

@inproceedings{zhang2025kgqagen,
  author = {Zhang, Liangliang and Jiang, Zhuorui and Chi, Hongliang and Chen, Haoyang and ElKoumy, Mohammed and Wang, Fali and Wu, Qiong and Zhou, Zhengyi and Pan, Shirui and Wang, Suhang and Ma, Yao},
  title = {Diagnosing and Addressing Pitfalls in {KG-RAG} Datasets: Toward More Reliable Benchmarking},
  booktitle = {Advances in Neural Information Processing Systems},
  volume = {38},
  year = {2025},
  pages = {75490--75527},
}

@inproceedings{lin2026hybridrag,
  author = {Lin, Junhong and Zhang, Bing and Wang, Song and Liu, Ziyan and Gutfreund, Dan and Shun, Julian and Zhu, Yada},
  title = {How Much Reasoning Do Retrieval-Augmented Models Add beyond {LLM}s? A Benchmarking Framework for Multi-Hop Inference over Hybrid Knowledge},
  booktitle = {Proceedings of the 32nd ACM SIGKDD Conference on Knowledge Discovery and Data Mining V.2},
  year = {2026},
  pages = {9403--9414},
  doi = {10.1145/3770855.3817481},
  publisher = {ACM},
}

@inproceedings{hu2020xtreme,
  author = {Hu, Junjie and Ruder, Sebastian and Siddhant, Aditya and Neubig, Graham and Firat, Orhan and Johnson, Melvin},
  title = {{XTREME}: A Massively Multilingual Multi-task Benchmark for Evaluating Cross-lingual Generalization},
  booktitle = {International Conference on Machine Learning},
  pages = {4411--4421},
  year = {2020},
}

@inproceedings{ahuja2023mega,
  author = {Ahuja, Kabir and Diddee, Harshita and
          Hada, Rishav and Ochieng, Millicent and
          Ramesh, Krithika and Jain, Prachi and
          Nambi, Akshay and Ganu, Tanuja and
          Segal, Sameer and Ahmed, Mohamed and
          Bali, Kalika and Sitaram, Sunayana},
  title = {{MEGA}: Multilingual Evaluation of Generative {AI}},
  booktitle = {Conference on Empirical Methods in Natural Language Processing},
  pages = {4232--4267},
  year = {2023},
  doi = {10.18653/v1/2023.emnlp-main.258},
  publisher = {Association for Computational Linguistics},
}

@article{wang2021wikidata5m,
  author = {Wang, Xiaozhi and Gao, Tianyu and Zhu, Zhaocheng and Zhang, Zhengyan and Liu, Zhiyuan and Li, Juanzi and Tang, Jian},
  title = {{KEPLER}: A Unified Model for Knowledge Embedding and Pre-trained Language Representation},
  journal = {Transactions of the Association for Computational Linguistics},
  volume = {9},
  pages = {176--194},
  year = {2021},
  doi = {10.1162/tacl_a_00360},
  publisher = {MIT Press},
}

@inproceedings{liu2025recoglab,
  author = {Liu, Andrew and Prior, Henry and Balasubramaniam, Gargi and Moroshko, Rivka and Zait, Amir and Labzovsky, Ilia and Karmon, Danny and Dasgupta, Ishita and Stachenfeld, Kim and Marino, Kenneth},
  title = {{ReCogLab}: A Framework Testing Relational Reasoning \& Cognitive Hypotheses on {LLM}s},
  booktitle = {The Thirteenth International Conference on Learning Representations},
  year = {2025},
}
\bibliographystyle{colm2026_conference}

\appendix
\section{Appendix}

\subsection{Model Checkpoint Inventory}
\label{app:model-inventory}

We evaluate 43 model checkpoints in total.

\paragraph{Qwen2.5 Base.}
Qwen2.5-0.5B, Qwen2.5-1.5B, Qwen2.5-3B, Qwen2.5-7B,
Qwen2.5-14B, and Qwen2.5-32B.

\paragraph{Qwen2.5 Instruct.}
Qwen2.5-0.5B-Instruct, Qwen2.5-1.5B-Instruct,
Qwen2.5-3B-Instruct, Qwen2.5-7B-Instruct,
Qwen2.5-14B-Instruct, and Qwen2.5-32B-Instruct.

\paragraph{Qwen3.}
Qwen3-0.6B, Qwen3-1.7B, Qwen3-4B, Qwen3-8B, and Qwen3-14B.

\paragraph{Qwen3.5.}
Qwen3.5-0.8B, Qwen3.5-2B, Qwen3.5-4B, Qwen3.5-9B,
and Qwen3.5-27B.

\paragraph{Gemma.}
Gemma~3 270M IT, Gemma~3 4B IT, Gemma~3 12B IT,
Gemma~3 27B IT, Gemma~4 E4B IT, Gemma~4 26B-A4B IT,
and Gemma~4 31B IT.

\paragraph{LLaMA.}
Llama~3.2 1B Instruct, Llama~3.2 3B Instruct,
Llama~3.1 8B Instruct, and Llama~3.1 70B Instruct.

\paragraph{Granite.}
Granite~3.3 2B Instruct, Granite~4.0 Micro,
Granite~4.0 H-Tiny, Granite~3.3 8B Instruct,
and Granite~4.0 H-Small.

\paragraph{DeepSeek.}
DeepSeek-R1-Distill-Qwen-1.5B,
DeepSeek-R1-Distill-Qwen-7B, and
DeepSeek-R1-Distill-Qwen-14B.

\paragraph{Gemini.}
Gemini~3 Flash and Gemini~3.1 Pro.

\subsection{Full Hypothesis Definitions and Statistical Procedures}
\label{app:hypotheses}

This appendix provides the complete mathematical definitions corresponding to the compact hypothesis descriptions in the main paper. For hypotheses involving model scale, $A_S$, $A_M$, and $A_L$ denote mean accuracy within the Small ($<7$B), Medium ($\geq7$B and $<14$B), and Large ($\geq14$B) buckets of a coherent model series.

\paragraph{H1: Diminishing Gains with Model Scale}

Let
\[
G_{SM}=A_M-A_S
\]
denote the gain from Small to Medium and
\[
G_{ML}=A_L-A_M
\]
the gain from Medium to Large. We define
\[
D_{\mathrm{scale}}
=
G_{SM}-G_{ML}.
\]

H1 is directionally supported when
\[
G_{SM}>0,
\qquad
G_{ML}>0,
\qquad
D_{\mathrm{scale}}>0.
\]

Thus, both scaling transitions must improve accuracy, and the Small-to-Medium gain must exceed the Medium-to-Large gain.

\paragraph{H2: Accuracy Increases with Model Size}

H2 tests the two adjacent scaling contrasts
\[
A_M>A_S
\]
and
\[
A_L>A_M.
\]

Support is reported separately by transition. A Medium result indicates that $A_M>A_S$, while a Large result indicates that $A_L>A_M$.

\paragraph{H3: Apply-to-Analyze Performance Drop}

We define
\[
D_{\mathrm{AA}}
=
A_{\mathrm{Apply}}
-
A_{\mathrm{Analyze}}.
\]

H3 tests
\[
D_{\mathrm{AA}}>0.
\]

In addition to directional and statistical support, we mark an Apply-to-Analyze ``cliff'' when the drop is statistically significant and the lower bound of its 95\% relation-cluster bootstrap interval exceeds 10 percentage points.

\paragraph{H4: Veracity Ordering}

H4 tests the ordering
\[
A_{\mathrm{Neutralized}}
>
A_{\mathrm{Factual}}
>
A_{\mathrm{Counterfactual}}.
\]

A size bucket supports H4 only when both component inequalities
\[
A_{\mathrm{Neutralized}}
>
A_{\mathrm{Factual}}
\]
and
\[
A_{\mathrm{Factual}}
>
A_{\mathrm{Counterfactual}}
\]
hold in the expected direction.

\paragraph{H5: Factual--Counterfactual Gap Widens with Cognitive Complexity}

For Bloom level $b$, define
\[
G_b
=
A_{\mathrm{Factual},b}
-
A_{\mathrm{Counterfactual},b}.
\]

We summarize lower-complexity levels as
\[
G_{\mathrm{low}}
=
\frac{
G_{\mathrm{Remember}}
+
G_{\mathrm{Understand}}
}{2},
\]
and higher-complexity levels as
\[
G_{\mathrm{high}}
=
\frac{
G_{\mathrm{Analyze}}
+
G_{\mathrm{Evaluate}}
}{2}.
\]

Apply is omitted because it occupies the middle position in the five-level sequence.

We define
\[
D_{\mathrm{FC}}
=
G_{\mathrm{high}}
-
G_{\mathrm{low}},
\]
and H5 tests
\[
D_{\mathrm{FC}}>0.
\]

\paragraph{H6: Relation Usage Frequency Has a Negligible Effect}

For each size bucket, let
\[
A_{\mathrm{Low}},
\qquad
A_{\mathrm{Medium}},
\qquad
A_{\mathrm{High}}
\]
denote mean accuracy on relations in the three usage-frequency tiers.

We define the observed spread as
\[
R_{\mathrm{usage}}
=
\max
\{
A_{\mathrm{Low}},
A_{\mathrm{Medium}},
A_{\mathrm{High}}
\}
-
\min
\{
A_{\mathrm{Low}},
A_{\mathrm{Medium}},
A_{\mathrm{High}}
\}.
\]

The effect is considered practically negligible when
\[
R_{\mathrm{usage}}<0.05,
\]
corresponding to a maximum spread below five percentage points.

For statistical equivalence, we consider the three pairwise contrasts:
\[
A_{\mathrm{Low}}-A_{\mathrm{Medium}},
\]
\[
A_{\mathrm{Low}}-A_{\mathrm{High}},
\]
and
\[
A_{\mathrm{Medium}}-A_{\mathrm{High}}.
\]

A bucket receives statistical support only when the corresponding 95\% relation-cluster bootstrap intervals lie entirely inside the $\pm5$ percentage-point equivalence region.

The confirmatory H6 claim concerns models at or above 7B parameters, so Medium and Large are the confirmatory buckets. Small is reported only as an exploratory diagnostic.

\paragraph{H7: Neutralized Advantage Is Concentrated at Higher Bloom Levels}

For each Bloom level $b$, define the Neutralized-over-Factual advantage
\[
N_b
=
A_{\mathrm{Neutralized},b}
-
A_{\mathrm{Factual},b}.
\]

We summarize the lower levels by
\[
N_{\mathrm{low}}
=
\frac{
N_{\mathrm{Remember}}
+
N_{\mathrm{Understand}}
}{2},
\]
and the higher levels by
\[
N_{\mathrm{high}}
=
\frac{
N_{\mathrm{Analyze}}
+
N_{\mathrm{Evaluate}}
}{2}.
\]

Apply is omitted as the middle level.

We define
\[
D_{\mathrm{NF}}
=
N_{\mathrm{high}}
-
N_{\mathrm{low}},
\]
and H7 tests
\[
D_{\mathrm{NF}}>0.
\]

Thus, H7 asks whether removing entity identity provides a larger relative benefit under higher cognitive demand.

\paragraph{H8: Evaluate Is a Recurring Bloom Bottleneck}

Let
\[
A_{\mathrm{other}}
=
\frac{
A_{\mathrm{Remember}}
+
A_{\mathrm{Understand}}
+
A_{\mathrm{Apply}}
+
A_{\mathrm{Analyze}}
}{4}.
\]

We define the Evaluate bottleneck gap as
\[
D_{\mathrm{Eval}}
=
A_{\mathrm{other}}
-
A_{\mathrm{Evaluate}}.
\]

H8 tests
\[
D_{\mathrm{Eval}}>0.
\]

This definition asks whether Evaluate is harder than the average of the other four Bloom levels rather than requiring it to be worse than each level individually.

\paragraph{H9: Item-Level Difficulty Is Stable Across Languages}

For item $i$ and language $\ell$, let $a_{i,\ell}$ denote its mean accuracy within a model series and size bucket. For every pair of languages $(\ell_1,\ell_2)$, we compute
\[
\rho^{\mathrm{item}}_{\ell_1,\ell_2}
=
\operatorname{Spearman}
\left(
\{a_{i,\ell_1}\}_i,
\{a_{i,\ell_2}\}_i
\right).
\]

With six languages, this yields 15 language-pair correlations. We summarize item-level cross-language stability by
\[
R_{\mathrm{item}}
=
\frac{1}{15}
\sum_{\ell_1<\ell_2}
\rho^{\mathrm{item}}_{\ell_1,\ell_2},
\]
and H9 tests
\[
R_{\mathrm{item}}>0.
\]

Significance is evaluated using a one-sided sign-flip test over the 15 pairwise Spearman correlations, testing whether their mean is greater than zero. Confidence intervals are obtained by bootstrap resampling the 15 language-pair correlation values and recomputing their mean.

\paragraph{H10: Relation-Level Difficulty Hierarchy Is Stable Across Languages}

For relation $r$ and language $\ell$, let $A_{r,\ell}$ denote mean accuracy on that relation. For every pair of languages, compute
\[
\rho^{\mathrm{rel}}_{\ell_1,\ell_2}
=
\operatorname{Spearman}
\left(
\{A_{r,\ell_1}\}_r,
\{A_{r,\ell_2}\}_r
\right).
\]

We summarize relation-level stability by
\[
R_{\mathrm{rel}}
=
\frac{1}{15}
\sum_{\ell_1<\ell_2}
\rho^{\mathrm{rel}}_{\ell_1,\ell_2}.
\]

H10 tests
\[
R_{\mathrm{rel}}>0.
\]

A positive result means that relations that are comparatively difficult in one language tend to retain their relative difficulty in the others.

\paragraph{Statistical Inference}

Questions derived from the same Wikidata relation share a source triple and therefore are not treated as independent observations.

For H1--H5, H7, and H8, we first construct the hypothesis-specific matched contrast and then average the contrast within each of the 30 source relations. Directional hypotheses are evaluated using one-sided relation-cluster sign-flip permutation tests with $10^5$ Monte Carlo permutations. We report 95\% relation-cluster bootstrap confidence intervals based on $2\times10^4$ resamples.

Holm correction is applied at $\alpha=0.05$ using hypothesis-specific correction families. For H1, the series-level $p$-values are corrected jointly across the available open-weight model series. For H2, the two adjacent size-transition tests are corrected within each series. For H4, the two component veracity contrasts are corrected jointly across the available size buckets within each open-weight series. For H3, H5, and H7--H10, the available S/M/L $p$-values are corrected within each open-weight series. For Gemini, H4 corrects its two component contrasts within each named model, whereas for H3, H5, and H7--H10 the two named-model $p$-values are corrected together.

H6 is an equivalence hypothesis rather than a directional hypothesis. We therefore use the predefined five-percentage-point equivalence region and assess whether the relation-cluster bootstrap intervals for the relevant pairwise usage-frequency contrasts lie entirely inside that region.

For H9, the statistic is the mean of the 15 pairwise item-level Spearman correlations. Significance is evaluated using a one-sided sign-flip test over the 15 correlation values, and confidence intervals are obtained by bootstrap resampling those language-pair correlations.

For H10, significance is evaluated using a relation-identity permutation test. Relation correspondences are independently shuffled across languages to break any stable cross-language relation ranking, after which the mean of the 15 pairwise Spearman correlations is recomputed. Confidence intervals are obtained by resampling relations and recomputing the full correlation statistic.

\subsection{Full Hypothesis Support Matrices}
\label{app:support-matrices}

Tables~\ref{tab:support-ft} and~\ref{tab:support-hy} give per-series and per-bucket support for all ten hypotheses; Table~\ref{tab:support-summary} in the main text summarizes them.

\begin{table}[h]
\caption{
Hypothesis support by model series in the \textbf{fully translated setting}.
Cells are shaded to convey strength of support at a glance:
\raisebox{-1pt}{\fcolorbox{black!25}{darkblue!62}{\rule{0pt}{6pt}\rule{10pt}{0pt}}}\ all present buckets significant \emph{and} an H3 cliff ($\dagger$);
\raisebox{-1pt}{\fcolorbox{black!25}{darkblue!45}{\rule{0pt}{6pt}\rule{10pt}{0pt}}}\ all present buckets significant (starred);
\raisebox{-1pt}{\fcolorbox{black!25}{darkblue!25}{\rule{0pt}{6pt}\rule{10pt}{0pt}}}\ partial significance (some buckets starred, some directional only);
\raisebox{-1pt}{\fcolorbox{black!25}{darkblue!10}{\rule{0pt}{6pt}\rule{10pt}{0pt}}}\ direction observed but no statistical support;
\raisebox{-1pt}{\fcolorbox{black!25}{red!15}{\rule{0pt}{6pt}\rule{10pt}{0pt}}}\ no support (\texttt{x});
\raisebox{-1pt}{\fcolorbox{black!25}{black!8}{\rule{0pt}{6pt}\rule{10pt}{0pt}}}\ not applicable (\textemdash{}).
Reading the cell contents: S, M, and L indicate that the required direction is observed in the Small, Medium, and Large model buckets, respectively; starred labels indicate statistical support under the hypothesis-specific procedure; Holm correction is applied where applicable.
$\dagger$ marks an H3 cliff whose lower 95\% relation-cluster bootstrap bound exceeds 10 percentage points.
H1 is a single series-level test rather than a per-bucket test.
Gemini models are reported individually because their parameter counts are undisclosed and are therefore not evaluated under H1 or H2; for Gemini, $*$ denotes statistical support and $\checkmark$ denotes the required direction without statistical support.
$\ddagger$ marks a value at the H6 five-percentage-point boundary under the rounded output.
}
\label{tab:support-ft}
\centering
\resizebox{\textwidth}{!}{%
\begin{tabular}{@{}lcccccccccc@{}}
\toprule
\textbf{Series}
& \textbf{H1}
& \textbf{H2}
& \textbf{H3}
& \textbf{H4}
& \textbf{H5}
& \textbf{H6}
& \textbf{H7}
& \textbf{H8}
& \textbf{H9}
& \textbf{H10} \\
\midrule

Qwen2.5 Base
& \hFull *
& \hPart S,M*,L*
& \hStrong S*,M*$\dagger$,L*$\dagger$
& \hWeak S,M,L
& \hWeak S,M
& \hWeak M
& \hFull S*,M*,L*
& \hFull S*,M*,L*
& \hFull S*,M*,L*
& \hFull S*,M*,L* \\

Qwen2.5 Instruct
& \hFull *
& \hPart S,M*,L*
& \hStrong S*,M*$\dagger$,L*$\dagger$
& \hPart S,M,L*
& \hWeak S
& \hNo \texttt{x}
& \hFull S*,M*,L*
& \hFull S*,M*,L*
& \hFull S*,M*,L*
& \hFull S*,M*,L* \\

Qwen3
& \hFull *
& \hPart S,M*,L*
& \hStrong S*,M*$\dagger$,L*$\dagger$
& \hWeak S,M,L
& \hPart S,M*,L
& \hWeak M,L
& \hPart S*,M,L
& \hFull S*,M*,L*
& \hFull S*,M*,L*
& \hFull S*,M*,L* \\

Qwen3.5
& \hFull *
& \hPart S,M*,L*
& \hStrong S*$\dagger$,M*$\dagger$,L*
& \hWeak S,M
& \hWeak S,L
& \hNo \texttt{x}
& \hPart S*,M
& \hFull S*,M*,L*
& \hFull S*,M*,L*
& \hFull S*,M*,L* \\

Gemma
& \hFull *
& \hPart S,M*,L*
& \hStrong S*$\dagger$,M*$\dagger$,L*
& \hWeak S,M
& \hWeak S,M,L
& \hNo \texttt{x}
& \hPart S*,M
& \hFull S*,M*,L*
& \hFull S*,M*,L*
& \hFull S*,M*,L* \\

LLaMA
& \hNo \texttt{x}
& \hPart S,M*,L*
& \hStrong S*,M*$\dagger$,L*$\dagger$
& \hWeak L
& \hNo \texttt{x}
& \hWeak S
& \hPart S*,M,L*
& \hPart S*,M,L*
& \hFull S*,M*,L*
& \hFull S*,M*,L* \\

Granite
& \hNo \texttt{x}
& \hPart S,M*,L*
& \hStrong S*,M*$\dagger$,L*$\dagger$
& \hWeak S,M,L
& \hWeak S,M
& \hNo \texttt{x}
& \hFull S*,M*,L*
& \hPart S*,M*,L
& \hFull S*,M*,L*
& \hFull S*,M*,L* \\

DeepSeek
& \hNo \texttt{x}
& \hPart S,M*
& \hPart S,M*,L*
& \hWeak M,L
& \hWeak S,L
& \hNo \texttt{x}
& \hPart S,M*,L
& \hFull S*,M*,L*
& \hFull S*,M*,L*
& \hFull S*,M*,L* \\

\midrule

Gemini~3 Flash
& \hNA \textemdash{}
& \hNA \textemdash{}
& \hNo \texttt{x}
& \hNo \texttt{x}
& \hFull *
& \hWeak $\checkmark^{\ddagger}$
& \hWeak \checkmark
& \hFull *
& \hFull *
& \hFull * \\

Gemini~3.1 Pro
& \hNA \textemdash{}
& \hNA \textemdash{}
& \hNo \texttt{x}
& \hNo \texttt{x}
& \hFull *
& \hWeak \checkmark
& \hNo \texttt{x}
& \hFull *
& \hFull *
& \hFull * \\

\bottomrule
\end{tabular}}
\end{table}

\begin{table}[h]
\caption{
Hypothesis support by model series in the \textbf{hybrid cross-lingual setting}.
Cell shading follows the same scale as Table~\ref{tab:support-ft} (deeper darkblue $=$ stronger support; rose $=$ no support). Gemini models were not evaluated in this setting.
}
\label{tab:support-hy}
\centering
\resizebox{\textwidth}{!}{%
\begin{tabular}{@{}lcccccccccc@{}}
\toprule
\textbf{Series}
& \textbf{H1}
& \textbf{H2}
& \textbf{H3}
& \textbf{H4}
& \textbf{H5}
& \textbf{H6}
& \textbf{H7}
& \textbf{H8}
& \textbf{H9}
& \textbf{H10} \\
\midrule

Qwen2.5 Base
& \hFull *
& \hPart S,M*,L*
& \hStrong S*$\dagger$,M*$\dagger$,L*$\dagger$
& \hWeak S,M,L
& \hNo \texttt{x}
& \hWeak M,L
& \hFull S*,M*,L*
& \hFull S*,M*,L*
& \hFull S*,M*,L*
& \hFull S*,M*,L* \\

Qwen2.5 Instruct
& \hFull *
& \hPart S,M*,L*
& \hStrong S*,M*$\dagger$,L*$\dagger$
& \hPart S,M,L*
& \hNo \texttt{x}
& \hWeak M
& \hFull S*,M*,L*
& \hFull S*,M*,L*
& \hFull S*,M*,L*
& \hFull S*,M*,L* \\

Qwen3
& \hFull *
& \hPart S,M*,L*
& \hStrong S*,M*$\dagger$,L*$\dagger$
& \hWeak S,M,L
& \hWeak S,M,L
& \hWeak M,L
& \hFull S*,M*,L*
& \hFull S*,M*,L*
& \hFull S*,M*,L*
& \hFull S*,M*,L* \\

Qwen3.5
& \hFull *
& \hPart S,M*,L*
& \hStrong S*$\dagger$,M*$\dagger$,L*
& \hPart S,M*,L
& \hPart S,L*
& \hNo \texttt{x}
& \hPart S*,M,L
& \hFull S*,M*,L*
& \hFull S*,M*,L*
& \hFull S*,M*,L* \\

Gemma
& \hFull *
& \hPart S,M*,L*
& \hStrong S*$\dagger$,M*$\dagger$,L*
& \hPart S*,M,L
& \hPart S,M,L*
& \hWeak L
& \hPart S*,M,L
& \hFull S*,M*,L*
& \hFull S*,M*,L*
& \hFull S*,M*,L* \\

LLaMA
& \hNo \texttt{x}
& \hPart S,M*,L*
& \hPart S,M*$\dagger$,L*$\dagger$
& \hPart L*
& \hWeak L
& \hWeak S,M
& \hFull S*,M*,L*
& \hPart S*,M,L*
& \hFull S*,M*,L*
& \hFull S*,M*,L* \\

Granite
& \hNo \texttt{x}
& \hPart S,M*,L*
& \hStrong S*$\dagger$,M*$\dagger$,L*$\dagger$
& \hPart S,M*,L
& \hWeak M
& \hWeak M
& \hFull S*,M*,L*
& \hPart S*,M*,L
& \hFull S*,M*,L*
& \hFull S*,M*,L* \\

DeepSeek
& \hFull *
& \hPart S,M*,L
& \hPart S,L*$\dagger$
& \hWeak M,L
& \hWeak L
& \hWeak S
& \hFull S*,M*,L*
& \hPart M*,L*
& \hFull S*,M*,L*
& \hFull S*,M*,L* \\

\bottomrule
\end{tabular}}
\end{table}

\subsection{Benchmark Construction Details}
\label{app:construction}

\paragraph{Relation candidate pool and usage tiers.}
We construct a candidate pool of Wikidata properties starting from the relations in Wikidata5M \citep{wang2021wikidata5m}. For each relation, we take its primary English alias as a human-readable handle, deduplicate, and randomly shuffle the resulting list using a fixed seed, retaining up to $N{=}1000$ candidates for downstream sampling. We estimate relation usage frequency by querying the MediaWiki search API with \texttt{haswbstatement:PID} and using the returned \texttt{totalhits} as a proxy. This measures the prevalence of the property in Wikidata-indexed content and should not be interpreted as a direct estimate of model pretraining exposure. We apply a $\log(1+\text{hits})$ transformation to reduce skew and split values into tertiles, defining Low, Medium, and High usage tiers. We then sample up to 30 properties per tier using a fixed random seed, yielding a candidate pool of up to 90 properties for manual curation.

\paragraph{Triple extraction and manual curation.}
For each sampled property $p$, we extract exactly one representative factual triple $(h,p,t)$ from popular English Wikipedia pages, ranked by sitelink count as a proxy for popularity. From the automatically sampled candidates, we manually curate a final set of 30 relations, 10 per usage tier, excluding relations involving sensitive topics, overly technical semantics, or cases where multiple relations feature the same entity. We prioritize semantic clarity and domain diversity across geography, science, engineering, sports, entertainment, and linguistics.

\subsection{Item Quality Control}
\label{app:quality-control}

All generated questions, answer options, and answer keys were manually reviewed by the authors for factual correctness, preservation of the intended relation, answer uniqueness, and distractor plausibility. As an additional quality-control step, every item was independently evaluated using Gemini and Grok. Each model received the fact sentence, question, answer options, and intended answer key and was asked to identify cases in which it disagreed with the provided answer. All flagged cases were manually inspected. When disagreement occurred, we examined whether it reflected model error or plausible ambiguity in the question or answer options. Valid objections led to revisions of wording or distractors, followed by re-evaluation before inclusion.
\subsection{Language-Specific Performance in the Fully Translated Setting}
\label{app:language-heatmaps}

\setcounter{figure}{0}
\renewcommand{\thefigure}{A\arabic{figure}}
\providecommand{\theHfigure}{\arabic{figure}}
\renewcommand{\theHfigure}{relbloomheatmap.\arabic{figure}}

\begin{figure}[p]
  \centering
  \includegraphics[width=\linewidth,height=0.86\textheight,keepaspectratio]{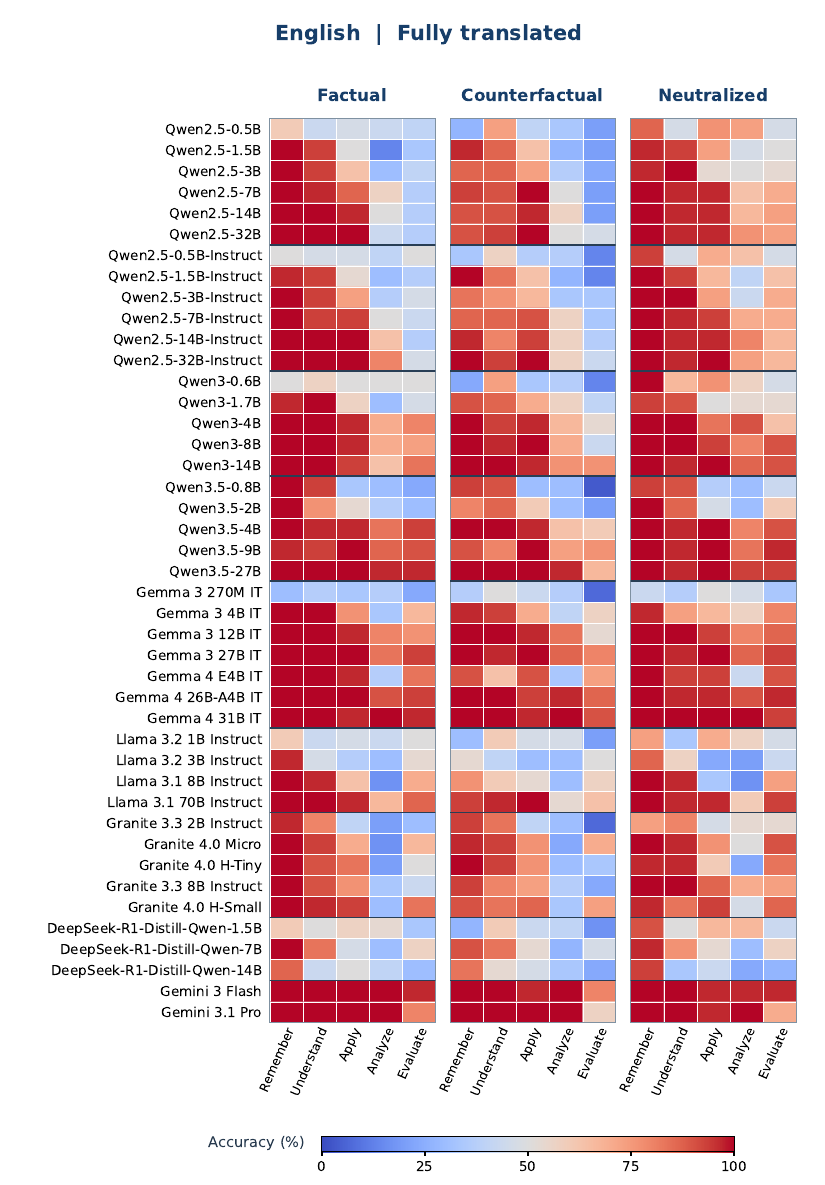}
  \caption{English: model accuracy by Bloom level and veracity condition in the fully translated setting.}
  \label{fig:heatmap-en}
\end{figure}

\begin{figure}[p]
  \centering
  \includegraphics[width=\linewidth,height=0.86\textheight,keepaspectratio]{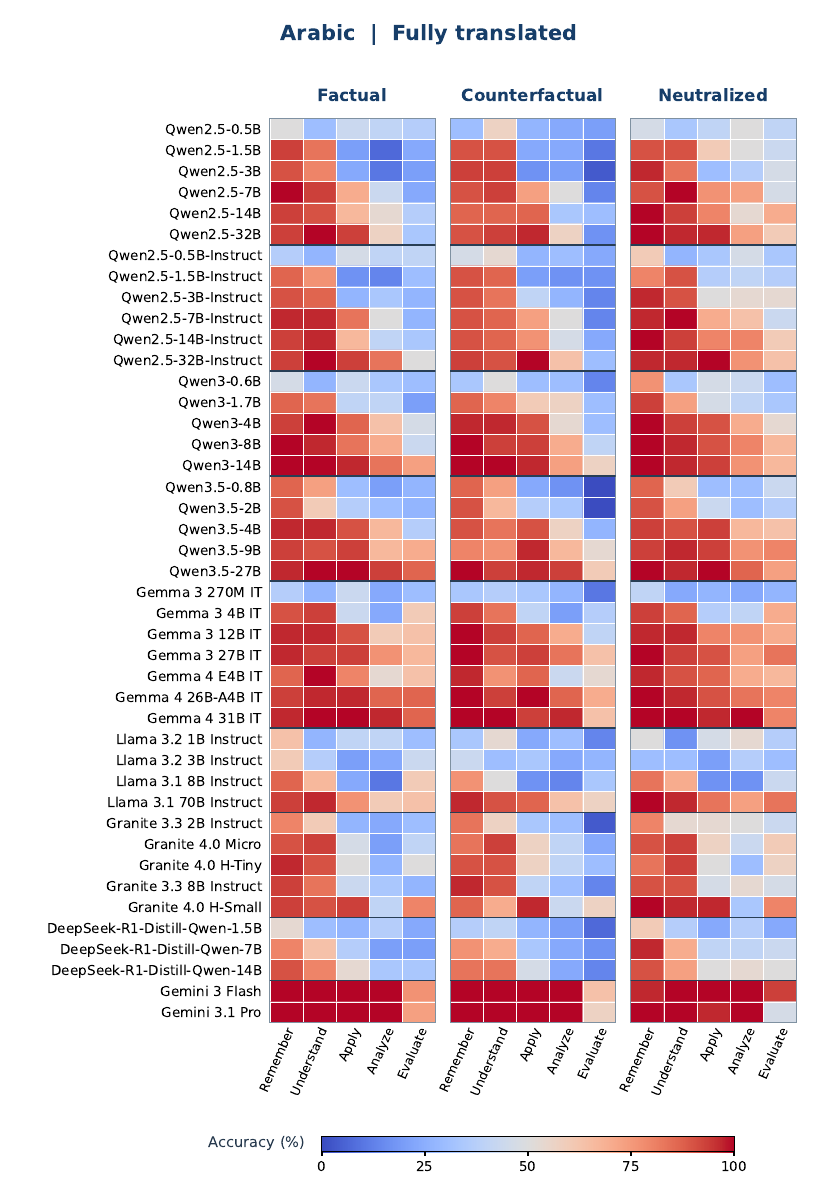}
  \caption{Arabic: model accuracy by Bloom level and veracity condition in the fully translated setting.}
  \label{fig:heatmap-ar}
\end{figure}

\begin{figure}[p]
  \centering
  \includegraphics[width=\linewidth,height=0.86\textheight,keepaspectratio]{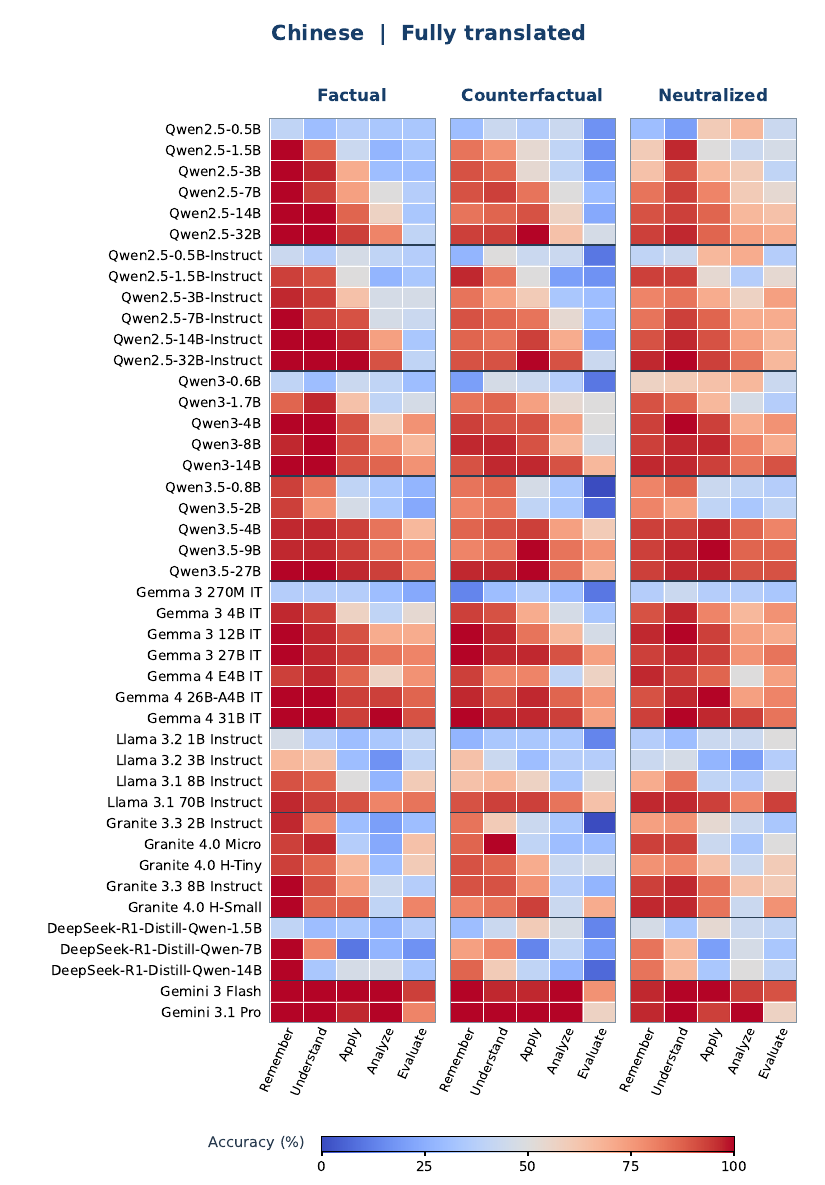}
  \caption{Chinese: model accuracy by Bloom level and veracity condition in the fully translated setting.}
  \label{fig:heatmap-zh}
\end{figure}

\begin{figure}[p]
  \centering
  \includegraphics[width=\linewidth,height=0.86\textheight,keepaspectratio]{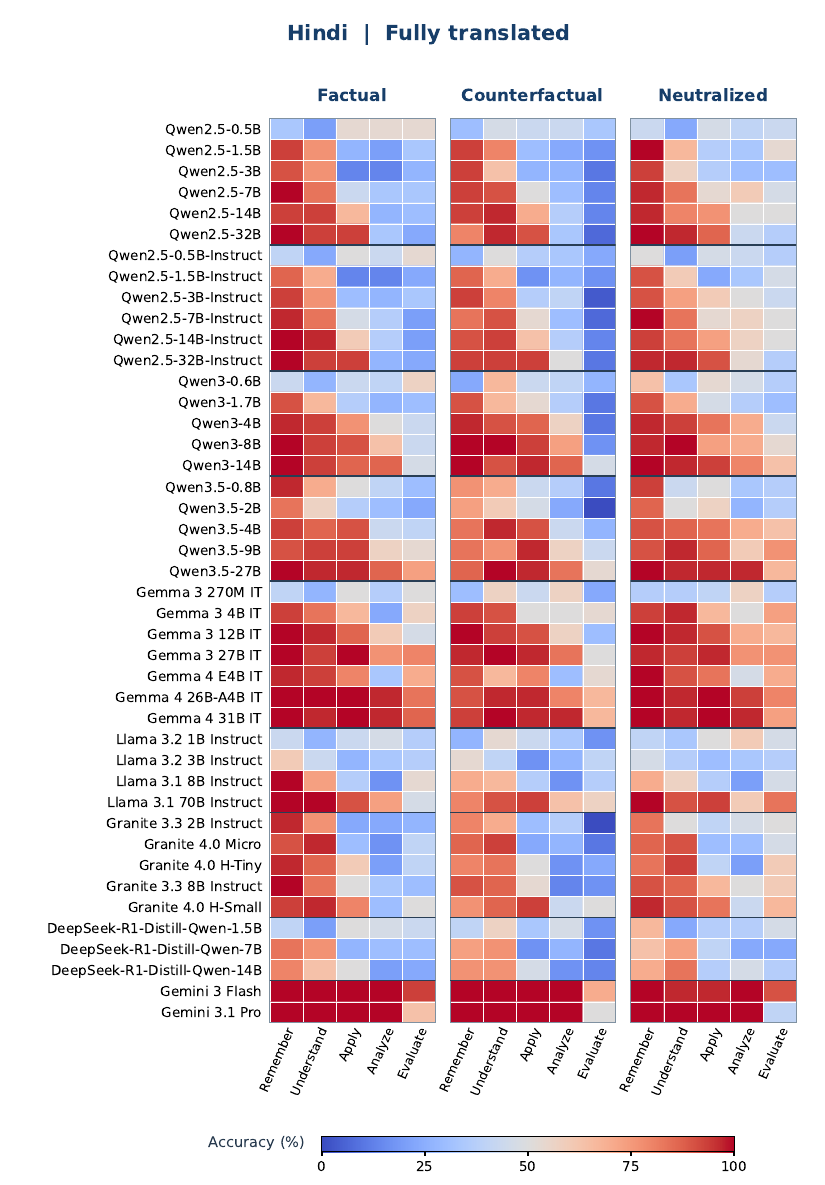}
  \caption{Hindi: model accuracy by Bloom level and veracity condition in the fully translated setting.}
  \label{fig:heatmap-hi}
\end{figure}

\begin{figure}[p]
  \centering
  \includegraphics[width=\linewidth,height=0.86\textheight,keepaspectratio]{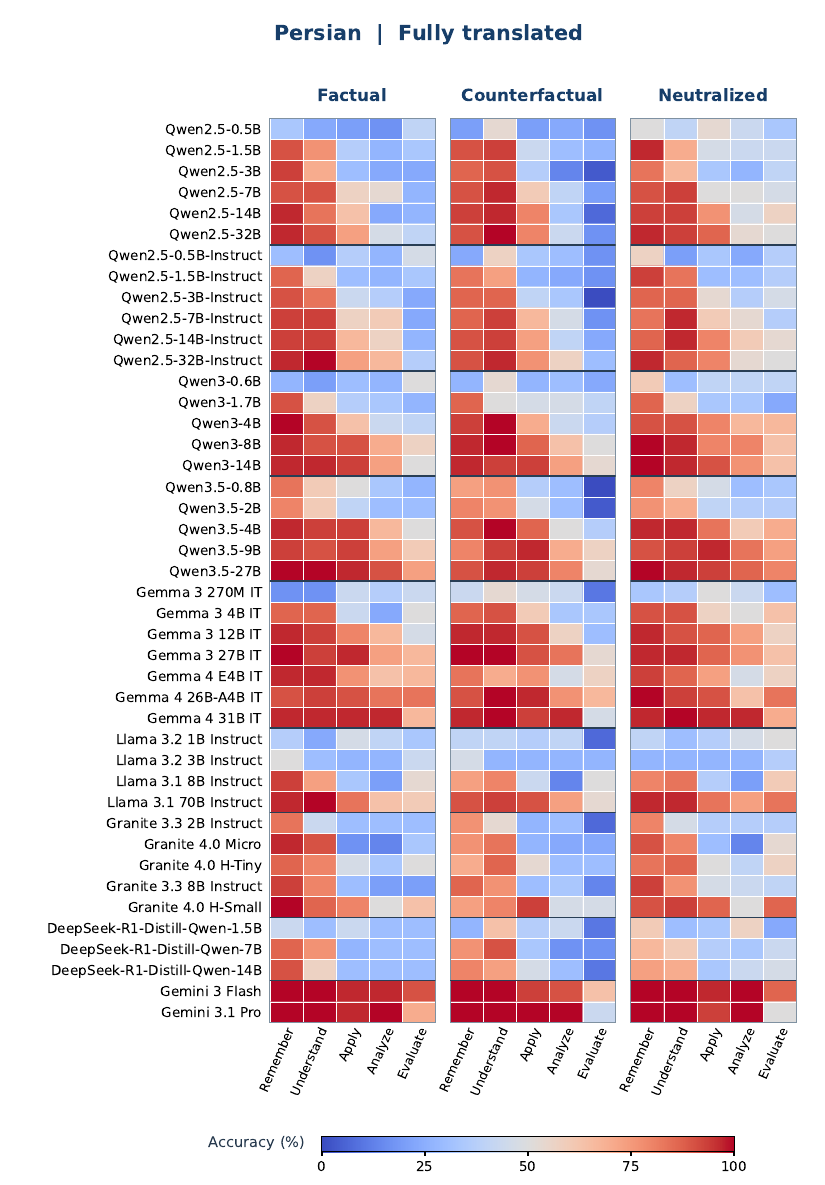}
  \caption{Persian: model accuracy by Bloom level and veracity condition in the fully translated setting.}
  \label{fig:heatmap-fa}
\end{figure}

\begin{figure}[p]
  \centering
  \includegraphics[width=\linewidth,height=0.86\textheight,keepaspectratio]{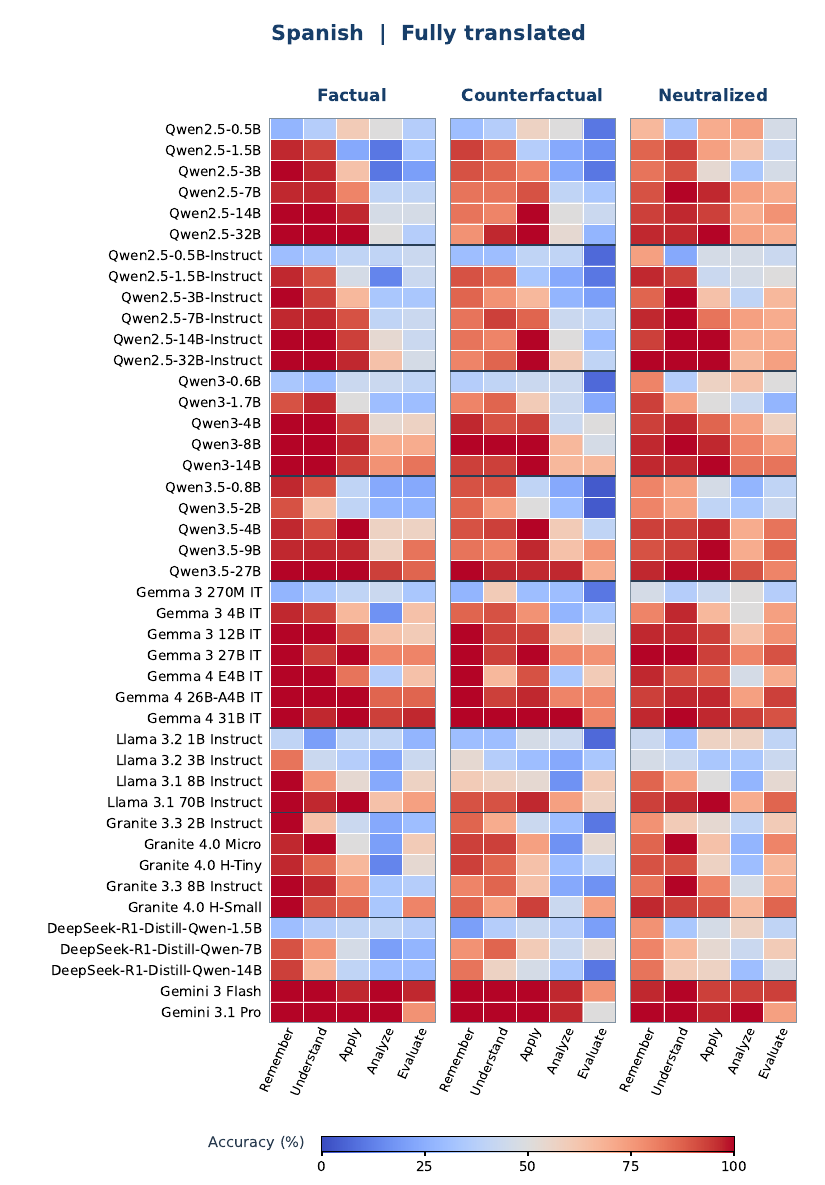}
  \caption{Spanish: model accuracy by Bloom level and veracity condition in the fully translated setting.}
  \label{fig:heatmap-es}
\end{figure}

Figures~\ref{fig:heatmap-en}--\ref{fig:heatmap-es} provide a language-specific detailed overview of model performance in the fully translated setting across English, Arabic, Chinese, Hindi, Persian, and Spanish. For each language, three heatmaps illustrate accuracy under the Factual, Counterfactual, and Neutralized conditions. The 43 model checkpoints (Appendix~\ref{app:model-inventory}) are arranged along the rows, while the columns represent the five Bloom-inspired cognitive levels: Remember, Understand, Apply, Analyze, and Evaluate. Each cell reports accuracy computed from 30 questions, with a consistent 0--100\% color scale applied across all 18 heatmaps to enable direct comparisons.

\clearpage
\end{document}